\pdfoutput=1

\documentclass[11pt]{article}

\usepackage[]{latex/acl}
\usepackage{tabularx, booktabs}
\usepackage{arydshln} 
\usepackage{times}
\usepackage{soul}
\usepackage{listings}
\usepackage{latexsym}
\usepackage{makecell}
\usepackage{graphicx}
\usepackage{multirow}
\usepackage{amsmath} 
\usepackage{amssymb} 
\usepackage{CJKutf8}
\usepackage{paralist}
\usepackage{xcolor}
\usepackage[T1]{fontenc}

\usepackage[utf8]{inputenc}

\usepackage{microtype}
\usepackage{inconsolata}
\usepackage{cleveref}
\usepackage{pifont}
\usepackage{array} 
\usepackage{color} 
\usepackage{float}    

\title{Agent-Pro: Learning to Evolve via Policy-Level Reflection and Optimization}

\author{Wenqi Zhang$^{1, *}$, Ke Tang$^{2,3,4,7, *}$, Hai Wu$^{2,3,5,7}$, Mengna Wang$^{2,6}$, Yongliang Shen$^{1}$\\
       {\bf Guiyang Hou$^{1}$, Zeqi Tan$^{1}$, Peng Li$^{2,3,7, \dagger}$, Yueting Zhuang$^{1}$, Weiming Lu$^{1, \dagger}$}\\
  $^1$College of Computer Science and Technology, Zhejiang University\\
  $^2$Institute of Software, Chinese Academy of Sciences  $^3$Nanjing Institute of Software Technology\\
  $^4$Nanjing University of Posts and Telecommunications\\
  $^5$Nanjing University of Information Science and Technology\\
  $^6$Beijing University of Technology $^7$University of Chinese Academy of Sciences, Nanjing \\
  \texttt{\{zhangwenqi, luwm\}@zju.edu.cn, lipeng@iscas.ac.cn}} 

\definecolor{darkgreen}{RGB}{0,100,0} 

\begin{document}
\maketitle
\renewcommand{\thefootnote}{\fnsymbol{footnote}} 
\footnotetext[1]{The first two authors have equal contributions. This work was done when the first author was an intern at Institute of Software, Chinese Academy of Sciences.}  
\footnotetext[2]{Corresponding author.}  
\renewcommand{\thefootnote}{\arabic{footnote}}

\begin{abstract}

Large Language Models (LLMs) exhibit robust problem-solving capabilities for diverse tasks. However, most LLM-based agents are designed as specific task solvers with sophisticated prompt engineering, rather than agents capable of learning and evolving through interactions. These task solvers necessitate manually crafted prompts to inform task rules and regulate LLM behaviors, inherently incapacitating to address complex dynamic scenarios e.g., large interactive games. In light of this, we propose \textbf{Agent-Pro}: an LLM-based \textbf{Agent} with \textbf{P}olicy-level \textbf{R}eflection and \textbf{O}ptimization that can learn a wealth of expertise from interactive experiences and progressively elevate its behavioral policy. Specifically, it involves a dynamic belief generation and reflection process for policy evolution. Rather than action-level reflection, Agent-Pro iteratively reflects on past trajectories and beliefs, "fine-tuning" its irrational beliefs for a better policy. Moreover, a depth-first search is employed for policy optimization, ensuring continual enhancement in policy payoffs. Agent-Pro is evaluated across two games: Blackjack and Texas Hold’em, outperforming vanilla LLM and specialized models. Our results show Agent-Pro can learn and evolve in complex and dynamic scenes, which also benefits numerous LLM-based applications \footnote{Code is in \url{https://github.com/zwq2018/Agent-Pro}}.









\end{abstract}

\begin{figure}[!t] 
    \centering
    \includegraphics[width=0.47\textwidth]{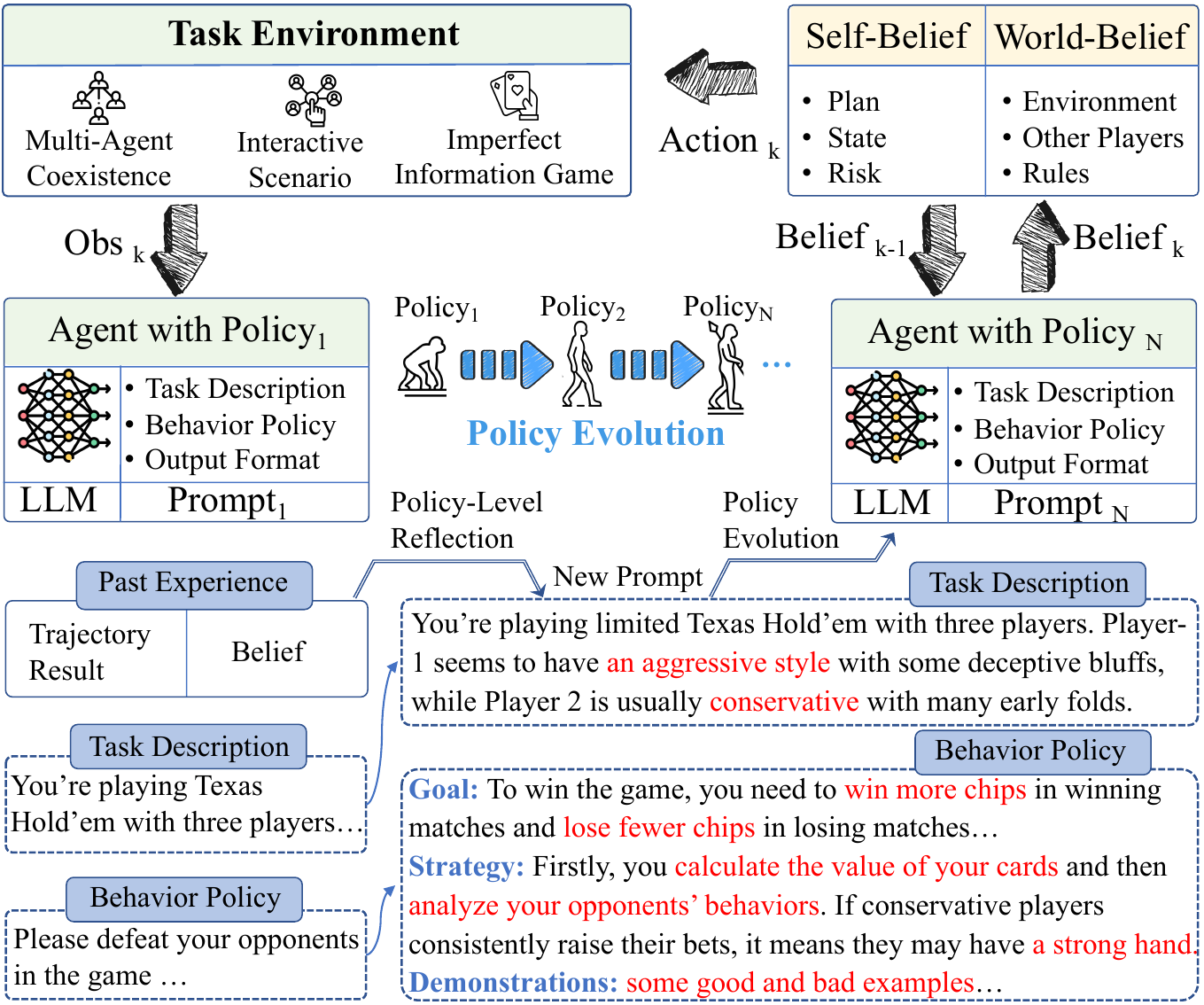}
    \caption{For interactive tasks, e.g., imperfect-information games, we propose a versatile agent framework capable of self-learning and evolving. Firstly, our agent constructs beliefs about itself and the environment. Then it autonomously updates its prompts through policy-level reflection on past trajectories and beliefs, evolving a better behavioral strategy.}
    \label{figure1}
\end{figure}


\section{Introduction}

Designing a human-level agent with robust problem-solving abilities has long been a vision in the academic community. This necessitates the agent to possess learning and generalization capabilities across a diverse array of tasks. The advent of Large Language Models (LLMs)~\citep{chowdhery2022palm, zhang2022opt, Zeng2023GLM, touvron2023llama, Chatgpt, gpt4, Touvron2023Llama2O} has shed light on this vision, especially they can be rapidly generalized across a wide range of tasks with only a few demonstrations~\citep{wei2022emergent, wei2022chain}. Benefiting from this, many systems built upon LLMs have showcased markedly enhanced performance such as question-answering~\citep{Yao2022ReActSR,schick2023toolformer,shen2023hugginggpt, Zhang2023DataCopilotBB,Madaan2023SelfRefineIR,zhang2022multi,zhang-etal-2023-expression}, code generation~\citep{Hong2023MetaGPTMP, Wu2023AutoGenEN}, and real-world application~\citep{qin2023toolllm, yang2023appagent}.


Despite these achievements, building a human-level agent remains a challenging endeavor. First, most LLM-based agents are designed for specific tasks through sophisticated prompts, including detailed task descriptions and behavioral specifications. However, numerous real-world tasks, e.g., business, company negotiations, and security, are more intricate with imperfect information, necessitating laborious efforts to design strategic behavior.


Second, most LLM-based agents do not consider interacting with task scenarios, and more critically, cannot learn from past experiences and evolve their behavioral strategies during interactions. In contrast, humans often learn and adjust their behaviors through interaction, especially in novel scenarios. In light of these, a promising yet under-explored topic emerges: \textbf{Can LLM-based agents learn and elevate behavioral strategies by interacting with the environment like humans}? It should be an indispensable ability of a human-level agent.


Recently, numerous studies~\citep{Shinn2023ReflexionAA,wang2023voyager,zhang2023proagent,zhao2023expel,qian2024investigate} undertake intriguing explorations, e.g., utilizing feedback for self-correction at the action-level. Besides, several efforts also explore deploying LLM in interactive games, including StarCraft~\citep{ma2023large}, Minecraft~\citep{wang2023voyager}, strategy-based gaming~\citep{Bakhtin2022HumanlevelPI, guo2023suspicion,xu2023exploring,xu2023language}. 

Similarly, we first evaluate LLM-based agents with the self-correction strategy in dynamic interactive scenarios, such as multi-player Texas Hold'em, which is a zero-sum game with imperfect information. However, we observe that it loses most of the rounds to its opponents, even the most advanced LLMs. Upon examining its reasoning thoughts and actions, we find that it often adopts irrational behaviors and is unable to deduce effective strategies from long action sequences. 



To answer the above question, the Theory of Mind (ToM)~\citep{premack1978does} may provide some insight. In this framework, each human develops perceptions of himself (self-belief) and the external environment (social-belief) in the social context, and then grounds their decisions on these beliefs, or adjusts incorrect beliefs in response to external feedback. Inspired by this, we advocate \textbf{Agent-Pro}: a LLM-based \textbf{Agent} with \textbf{P}olicy-level \textbf{R}eflection and \textbf{O}ptimization. Agent-Pro is endowed with the capacity to learn and evolve within environments, i.e., autonomously reflect on past experiences, calibrate its beliefs about itself and the environment, and optimize its behavior policy without parameter tuning.


%

Concretely, as shown in \Cref{figure1}, an LLM-based agent involves an LLM as the foundational model and some instructions in the prompt to regulate its behavior (policy). Upon observing partial information from the scenarios, Agent-Pro first updates its self-belief and world-belief, then makes decisions based on these beliefs. After exploring tasks, Agent-Pro performs a policy-level reflection and optimization on past trajectories, beliefs, and results. It autonomously "fine-tunes" its beliefs, searches for useful prompt instructions, and consolidates them into a new behavior policy.

The experiments in two zero-sum games, Blackjack and Texas Hold'em, demonstrate that Agent-Pro, after evolution, can defeat vanilla LLMs and specialized models, improving the game's payoffs. It indicates that Agent-Pro enhances its capabilities through interaction and reflection without human guidance. As depicted in~\Cref{figure1}, the initial prompt is quite simple (Left Bottom), but after learning and evolution, the Agent-Pro generates many practical instructions (Right Bottom). For instance, Agent-Pro records estimations of each opponent's style in \texttt{Task Description} and adds specific \texttt{Goals, Strategies} in \texttt{Behavior Policy}.

Our Agent-Pro is different from previous strategies, like Reflexion~\citep{Shinn2023ReflexionAA}. Firstly, Policy-level reflection is designed for policy updating in long-horizon tasks. It is aimed at long-horizon policy updating rather than immediate action correction. The input is a sequence of actions and delayed feedback, while the output is an optimized strategy, rather than a specific action. Therefore, policy-level reflection corrects irrational beliefs and optimizes the old policy into the new one. As introduced in~\Cref{policy-level reflection}, our policy-level reflection includes belief calibration, policy updates by refining behavioral guidelines and world modeling, and policy verification.

Besides, we innovatively distill long-term memory into Behavioral Guidelines and World Models through prompt optimization. Most previous strategies store historical experience as verbal long-term memory and use it for text-based reasoning. In contrast, we further construct an optimizable policy from long-term interactions, i.e., Behavioral Guidelines and Environmental Models. This includes self-summarized game objectives and rules, effective strategies derived from reflection, and demonstrative trajectories. The contributions of our work are as follows:
\begin{itemize}
\item We introduce Agent-Pro, a framework capable of learning and evolving within interactive games, empowering LLM-based agents to efficiently adapt to more complex dynamic tasks.

\item We devise a belief-aware decision-making process with self and world-belief, enhancing its capabilities for intricate tasks, i.e., generating more rational actions in interactive scenarios.

\item We utilize policy-level reflection and optimization to iteratively update prompt instructions, which empower Agent-Pro to progressively evolve from a novice to a skilled veteran with many strategic behaviors.

\item After learning, Agent-Pro is evaluated in multiplayer games and defeats specialized models, gaining notable progress. It develops strategic skills like humans, e.g., actively cutting losses, bluffing, or disguising to influence others.
\end{itemize}

Not just in card games, similar scenarios abound in the real world as well. Through self-learning and evolution, Agent-Pro can enhance deployment effectiveness in those scenarios, expanding the capability boundaries of LLM-based agents notably.






%
\section{Problem Definition}
Our study focuses on multi-player imperfect information interactive games, with two characteristics:

\textbf{Imperfect Information.}\label{imperfect information games} Unlike perfect information games (e.g., chess), imperfect information scenarios are characterized by agents only having access to their own states and public information, without knowing the states of others, e.g., in Texas Hold'em, players cannot observe others’ cards, which is dissimilar to many LLM-based tasks.

\textbf{Dynamic Interaction.} There may be multiple agents in the environment, and they may influence each other. That is, the actions of one agent may lead to changes in the environment, which are unpredictable for other agents.

In real-world contexts, such as competition, company negotiations, and security, these scenarios can often be abstracted as multi-agent interactive scenarios with imperfect information. Research on this can offer viable solutions to many real-world problems. We select two games as our testbed: Blackjack and Limit Texas Hold'em with multi-player. Please refer to~\Cref{introduction of two games} for details.

\begin{figure*}[!t] 
    \centering
    \includegraphics[width=1\textwidth]{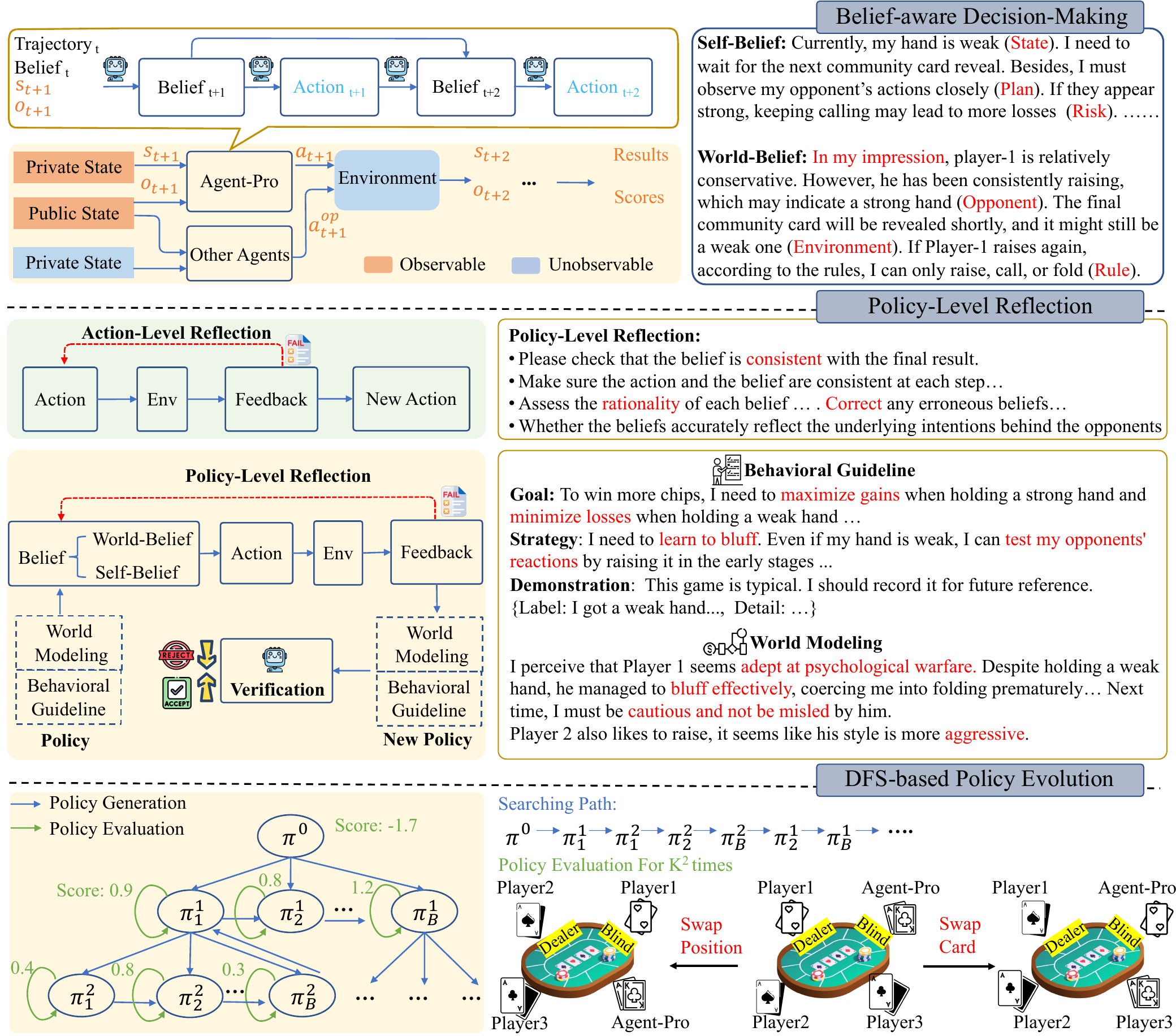}
    \caption{In a competitive multiplayer game with imperfect information, Agent-Pro designs a dynamic belief to enhance decision-making capabilities. It first updates its beliefs about the world and itself, then generates more coherent actions. To achieve policy-level reflection, Agent-Pro examines the beliefs associated with failed trajectories. It then summarizes prompt instructions, including World Modeling and Behavioral Guideline to calibrate incorrect beliefs. Lastly, Agent-Pro employs a DFS-based search to incrementally enhance policy effectiveness.}
    \label{figure2}
\end{figure*}

\section{Methods}
To empower agents to learn in interactive contexts, a typical method is reinforcement learning~\citep{Zhang2021LearningTN,ijcai2022p654}. This involves exploring highly rewarding actions through trial and error and solidifying these experiences into model parameters. Nonetheless, the training overhead for LLMs is substantial. Therefore, we employ a gradient-free "exploration-learning" strategy that enables LLM-based agents to learn through in-context learning. Specifically, we convert the policy learning into a prompt optimization process, i.e., LLM autonomously reflects and updates the prompt's instructions based on its exploration experience, solidifying the high-reward strategies into the prompts. Benefiting from LLM's generalization capabilities, our agent can summarize rules and learn specialized skills from a small number of samples like humans, making it well-suited for many real-world scenarios.


As shown in \Cref{figure2}, Agent-Pro comprises three components: (1) A Belief-Aware Decision-Making process. It first updates beliefs about the world and itself, rendering more coherent and consistent decisions in dynamic and imperfect game scenarios. (2) A Policy-Level Reflection. Rather than reflecting on a single action, our design empowers LLMs to self-reflect on irrational beliefs from failed experiences. Then, it summarizes these erroneous beliefs into specific prompt instructions, like acting strategy (Behavioral Guideline), descriptions of the task world, and conjectures about other players (World Modeling), etc, which can calibrate its incorrect beliefs, evolving into a better policy. (3) A Prompt Optimization process ensures that the agent's policy evolves for a higher payoff following a DFS-based search.

\subsection{Belief-aware Decision-Making Process}  \label{belief-aware process}
To develop an LLM-based agent better suited for interactive environments, we draw inspiration from the Theory of Mind (ToM)~\citep{premack1978does,li-etal-2023-theory, guo2023suspicion}. In this framework, human condenses perceptions of themselves (self-belief) and the external environment (social-belief) and then ground their decisions on these beliefs, or adjust incorrect beliefs in response to external feedback. We also design a belief-aware decision-making process for Agent-Pro, simulating human cognitive processes in social contexts.

First, we need to define the policy of an LLM-based agent, which refers to a specific behavioral strategy guiding the agent to interact and complete tasks. It often involves complex prompts designed by experts, covering task rules, strategies, and output formats. In a zero-sum game with $K\text{+}1$ players (assuming playing order is $(op_1, our, op_2,.., op_K)$), we denote the policy of our agent as $\pi$ with some observable information, containing agent's private information $\textcolor{red}{s_{t}}$, public information $\textcolor{blue}{o_{t}}$, our own action $\textcolor{cyan}{a_{t}}$, and the actions of all opponents $a^{op_{1}}_{t}, a^{op_{2}}_{t}, .., a^{op_{K}}_{t}$, where $t$ means $t$-th rounds of a game. Therefore a complete game trajectory spanning $t$ rounds: 
\begin{equation}
\begin{aligned}
\mathcal{H}_{0:t} = \{ &(\textcolor{red}{s_{0}}, \textcolor{blue}{o_{0}}, a^{op_{1}}_{0}, \textcolor{cyan}{a_{0}}, a^{op_{2}}_{0},...,a^{op_{K}}_{0}),\\
&\quad \quad \quad \quad \quad \vdots\\
&(\textcolor{red}{s_{t}}, \textcolor{blue}{o_{t}},a^{op_{1}}_{t}, \textcolor{cyan}{a_{t}},a^{op_{2}}_{t},...,a^{op_{K}}_{t})\}
\end{aligned}
\end{equation}

As shown in \Cref{figure2}, when making a decision, Agent-Pro first generates a dynamic belief $\textcolor{purple}{\xi}$ about itself (self-belief) and opponents (world-belief) in natural language. Then, it predicts an action based on the latest beliefs. For instance, for Texas Hold'em, Agent-Pro's understanding of its hand cards, plan, and potential risk constitutes its self-belief, while the conjectures about the opponents form its world-belief. These beliefs are updated in each decision-making cycle. Equipped with this, Agent-Pro can generate more coherent and consistent actions: 
\begin{equation}
\begin{aligned}
\textcolor{purple}{\xi_{t+1}}, \textcolor{cyan}{a_{t+1}}\sim\pi(\mathcal{H}_{0:t}, \textcolor{red}{s_{t+1}}, \textcolor{blue}{o_{t+1}},a^{op_{1}}_{t+1},\textcolor{purple}{\xi_{t}})
\end{aligned}
\end{equation}

When a game is over, we acquire the observable state $R$ (e.g., private hand cards after showdown) and the final scores $S$ of all players. The objective is to find an optimal $\pi^*$ to maximize $S(our)$.
\subsection{Policy-Level Reflection}  \label{policy-level reflection}
Equipped with an initial policy (a simple prompt) and a dynamic belief, Agent-Pro already possesses basic capabilities for game exploration. To further enhance Agent-Pro's capabilities, we design a learning mechanism via a policy-level reflection. 


Specifically, many text-based tasks have employed reflection strategies and immediate environmental feedback to correct prior actions. However, in many typical interaction scenarios with longer decision-making processes, action-level reflections are not directly applicable due to delayed feedback. Therefore, for such a long-horizon interaction process, Agent-Pro is instructed to focus on the rationality of beliefs and underlying behavioral policies rather than individual actions.





\lstset{
    framesep=20pt,
    rulesep=10pt,
    backgroundcolor=\color[RGB]{245,245,244},
    breaklines=true,
    breakindent=0pt,
    basicstyle=\ttfamily\small,
    escapeinside={(*@}{@*)}, 
}

\textbf{Belief Calibration} As depicted in~\Cref{figure2}, under the guidance of the current behavior policy, Agent-Pro generates actions based on self-belief and world-belief. If these beliefs are inaccurate, they may lead to irrational actions and eventual failure. Therefore, Agent-Pro examines the rationality of these beliefs based on the final results and reflectss on the reasons for the final failure. 
\begin{lstlisting}
(*@\textbf{Correctness}@*): Whether its beliefs about itself, the game, and its opponents align with the final results.
(*@\textbf{Consistency}@*): Whether each belief and action is self-contradictory.
(*@\textbf{Rationality}@*): Whether the beliefs accurately reflect the underlying intentions behind the opponents. 
(*@\textbf{Reasons}@*): Reflect on why it lost to its opponents, which beliefs are problematic, and what the underlying reasons are.
\end{lstlisting}

Lastly, to calibrate the incorrect beliefs, Agent-Pro summarizes these reflections and analyses about itself and the external world into specific instructions: \textbf{Behavioral Guideline} and \textbf{World Modeling}, where the former represents generalized behavioral strategies for this task, and the latter signifies its understanding and conjectures about the game world. For instance, in Texas Hold'em, Agent-Pro summarizes the following contents: 
\begin{lstlisting}
(*@\textbf{Behavioral Guideline}@*)
1-Please summarize a detailed goal based on your reflection on beliefs. {(*@\color{blue}{Goal}@*)}
2-What strategy helps you build correct belief and win at similar.. {(*@\color{blue}{Strategy}@*)}
3-Can this game be considered a typical example for future... {(*@\color{blue}{Demonstration}@*)} 
(*@\textbf{World Modeling}@*)
1-Accurately model each player to help build more precise beliefs about them, including action, and style.{(*@\color{blue}{Opponent}@*)} 
2-Describe any game rules or details that are easy to overlook...{(*@\color{blue}{Rule}@*)}
\end{lstlisting}

\noindent Agent-Pro summarizes high-level strategies within the \texttt{Behavioral Guideline} and describes the task and opponents in \texttt{World Modeling}. These instructions can calibrate previous incorrect beliefs and improve policy performance. The entire process can be formalized as follows:
\begin{equation}
\begin{aligned}
\text{Instruction}^{n+1}\!&\leftarrow\!\text{LLM}\left(\mathcal{H}^n_{0:T},\!\{\textcolor{purple}{\xi^n_1,\xi^n_2}..\},\!R^n,\!S^n\right)
\end{aligned} \label{policy update}
\end{equation}
where $\mathcal{H}^n_{0:T}$ denotes a complete trajectory at the $n$-th match, $\{\textcolor{purple}{\xi_1^n},\textcolor{purple}{\xi_2^n},...\}$ denotes the belief sequence, $R^n$ and $S^n$ means the final results and score. $\text{Instruction}^{n+1}$ denotes new generated \texttt{Behavioral Guideline} and \texttt{World Modeling}.


\textbf{Verification} After extracting these instructions, Agent-Pro verifies its efficacy. Agent-Pro incorporates these generated \texttt{Behavioral Guideline} and \texttt{World Modeling} into the prompt and then replays the same game again, i.e., the same opponents and initial conditions. If the final score improves, we retain them in the prompt. Otherwise, we regenerate a new one. If it fails to pass verification after three retries, we discard this trajectory $\mathcal{H}^{n}$:
\begin{equation}
\begin{aligned}
\pi^{n+1}\!&\!\xleftarrow{Verify}\!\pi^{n} \cup \text{Instruction}^{n+1}
\end{aligned} \label{policy verify}
\end{equation}
where $\cup$ means incorporates new instructions into the previous prompt for $\pi^{n+1}$. This new policy encompasses more effective instructions, empowering Agent-Pro to establish accurate self- and world beliefs and generate more rational actions. 

\subsection{DFS-based Policy Evolution}   \label{policy optimization}
To iteratively update the policy, we devise a policy optimization process based on depth-first search (DFS). It encompasses a policy evaluation process to assess the generalization ability of the new policy in novel game scenarios and a search mechanism to progressively find a better policy.

\textbf{Policy Evaluation} Each time the policy is updated, Agent-Pro is required to evaluate the new strategies. This evaluation process is distinct from the previous \texttt{Verification} step, as the \texttt{Verification} repeatedly utilizes the "training" data for evaluation and can not ensure the generalizability of the new policy. Hence, Agent-Pro conducts a thorough assessment of the new policy in novel trajectories. Besides, it is imperative to eliminate the influence of random factors when policy evaluation, e.g., a poor initial hand due to bad luck or an unfavorable playing order.



Therefore, we first randomly generate a new game for $K\text{+}1$ players. Then we sequentially swap both the hand cards and the playing order of each player, generating a total of $(K\text{+}1)^2$ combinations. To eliminate randomness, we concurrently use these $(K\text{+}1)^2$ games to evaluate Agent-Pro's new policy. We calculate the average score over the $(K\text{+}1)^2$ games for each player. Since the influences of hand-card quality and playing order are mitigated, the average score of all combinations can represent the true capabilities of each player. Lastly, we calculate the evaluating metrics:
\begin{equation}
\begin{aligned}
\Delta\!=\!\frac{1}{(K\text{+}1)^{2}}\!\!\sum_{j}^{(K+1)^2}\!\!\!\left[\mathcal{S}_{j}(\text{our})\!-\!\max_{i}\mathcal{S}_{j}(\text{op}_i)\right]\!
\end{aligned}
\end{equation}
where $i \in \{1, \ldots, K\}$ denotes the index of an opponent, and $j$ denotes the index of the games within $(K\text{+}1)^2$ combinations. The $\Delta$ assesses its gains relative to the strongest opponent, providing a comprehensive evaluation in multiplayer gaming scenarios.


\textbf{Policy Search}
Inevitably, sometimes the new policy does not bring an improvement in $\Delta$ in the new scenario. In such cases, we employ DFS to search for a better policy from other branches (i.e., other candidate policies). As shown in~\Cref{figure2}, when updating old policy $\pi^n$, we generate $B$ candidate policies $\{\pi^{n+1}_1$, $\pi^{n+1}_2$,..., $\pi^{n+1}_B\}$, forming $B$ branches. Then, we first calculate $\Delta^{n+1}_{1}$ for new policy $\pi^{n+1}_1$ and compare it with $\Delta^n$. If $\Delta^{n+1}_{1}$ is greater than $\Delta^n$, we accept this evolutionary. Otherwise, we reject $\pi^{n+1}_1$ and consider $\pi^{n+1}_2$. If none of the $B$ candidate policies $\pi^{n+1}$ enhance Agent-Pro's performance, we backtrack to $\pi^{n}$ and consider its sibling nodes $\pi^{n}_2$. Similarly, Agent-Pro explores the environment using $\pi^{n}_2$, then also updates $B$ candidate policies and searches in a depth-first manner. Ultimately, we select the policy with the highest $\Delta$ across the entire policy tree.


\section{Game: Blackjack} \label{exp: blackjack}
\textbf{Environment Settings} We employ the RLCard~\citep{zha2019rlcard} as our simulators for two games. We train two reinforcement learning agents as opponents: DQN~\citep{mnih2015human}, and Deep Monte Carlo Search (DMC)~\citep{zha2021douzero}. Please refer to~\Cref{experiments details} for more details.


\subsection{Results} 
As shown in~\Cref{result of blackjack}, we report the win rates of each agent against the dealer over 900 games. We also provide the results of RL-based models and a human player in~\Cref{other result of blackjack} for reference. 

\textbf{Agent-Pro Significantly Surpasses the Baseline Agents Across most LLMs.} The results show that Agent-Pro significantly surpasses most baseline agents with an average advantage of +4\%. For example, On Qwen-72B and Llama2-70B, Agent-Pro significantly surpasses Reflexion with increases of +3.9\% and +11\%, respectively. For GPT-4, Blackjack is relatively simple, so the win rates of different strategies are quite similar. 

\begin{table}[t!]\small
    \centering
    \setlength\tabcolsep{1.5pt} 

    \begin{tabular}{ l c c c c}
    \toprule[1pt]
    \textbf{Win Rate $\uparrow$(\%)} & \multicolumn{4}{c}{\textbf{Based Models}}  \\ \hline
    \textbf{Strategy} & \textbf{Qwen-72B} & \textbf{Llama2-70B} & \textbf{GPT3.5}  &\textbf{GPT4}  \\
    Vanilla LLM     &  0.5  & 0.3  &  27.9 & 34    \\ 
    Radical LLM     &  0.6  & 0.4  &  1.8  & 11.5  \\ 
    ReAct           &  30.9 & 11.8 &  36.6 & \textbf{40.9}  \\ 
    Reflexion       &  32.3 & 12.1 &  36.7 & 40.8  \\ 
    Agent-Pro       &  \textbf{36.2} \tiny{$\uparrow $3.9} & \textbf{23.1} \tiny{$\uparrow $11.0} &  \textbf{38.2} \tiny{$\uparrow $1.5} & 40.4 \tiny{$\downarrow $0.5} \\
    - \texttt{w/o Learning}  &  34.1 & 8.0  &  37.4 & 40.6    \\
    \bottomrule[1pt]
    \end{tabular}
    \caption{All agents compete independently against the dealer and then we calculate their win rates. \texttt{w/o} means only with belief-aware decision-making process. $\uparrow$ shows the difference compared to the best baseline.}\label{result of blackjack}
\end{table}

\textbf{What has Agent-Pro learned from evolution?} Compared to ReAct and Reflexion, Agent-Pro is more robust. We find that this is due to the effective behavioral guidelines summarized by policy-level reflection. For instance, Agent-Pro summarizes two instructions as follows: \texttt{1-When you have achieved a relatively stable total hand value, choosing not to take risks is a good decision. 2-Analyze the dealer cards in World-belief,..., excessive risk-taking can lead to unfavorable outcomes..}. These self-summarized instructions can alert Agent-Pro to the risks associated with action \texttt{Hit}, thus making more rational decisions.

\begin{figure}[!t] 
    \centering
    \includegraphics[width=0.46\textwidth]{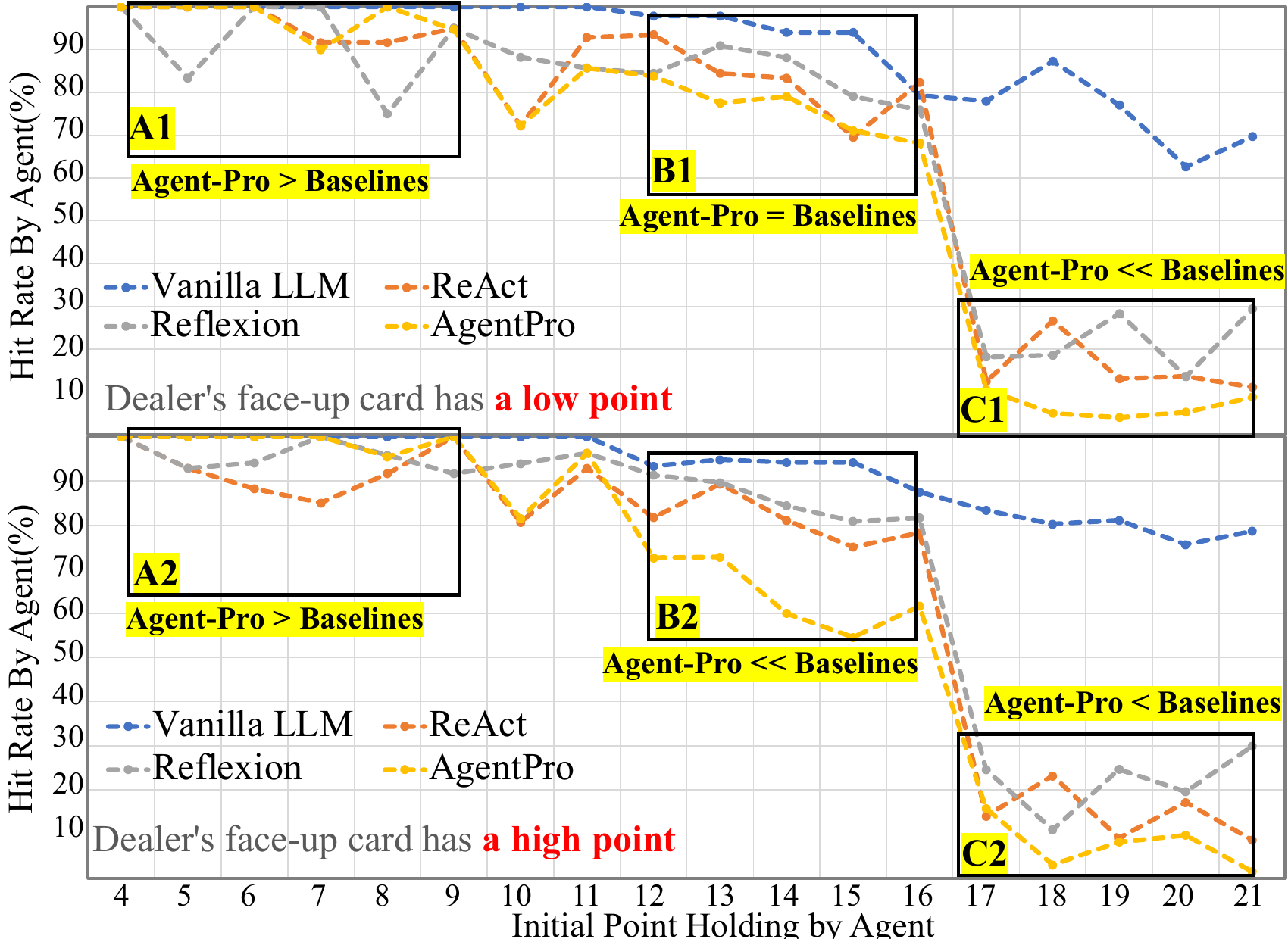}
    \caption{We analyze the hit rates of the agents under different initial point totals. Upper Figure: The dealer's face-up card has a low point. Lower figure: The dealer's face-up card has a high point.}
    \label{fig: hit rate}
\end{figure}

\subsection{Analyisis}
\textbf{Agent-Pro is More Rational than Baselines.} We further analyze the \texttt{Hit} rates of the agents under different initial point totals, i.e., the sum of the initial two cards. The hit rate represents whether the agent is willing to take risks to draw cards. At this point, the player needs to consider both their own hand and the dealer's hand to decide whether to take the risk. However, in~\Cref{fig: hit rate}, we observe that the baseline seems to only focus on its own hand, with no significant difference in behavior when the dealer's cards are high or low, whereas Agent-Pro is much more reasonable. For instance, for Agent-Pro, areas \texttt{B1} and \texttt{B2} show a clear difference. It tends to \texttt{Stand} when the dealer has high cards and \texttt{Hit} when the dealer has low cards. Because it believes the dealer is more likely to bust with high cards, making it not worth the risk for itself. We provide some detailed cases and evolution processes in~\Cref{Blackjack Case1,Blackjack Case2,Blackjack Case3,Blackjack Case4} to show their difference.




\begin{table*}[h!]\small
\centering
\setlength\tabcolsep{2pt} 
\begin{tabular}{l|llll|llll|llll}
\toprule[1pt]
\multirow{3}*{Agent Strategy} & \multicolumn{4}{c|}{Based Model = GPT3.5 } & \multicolumn{4}{c|}{Based Model = GPT4} &\multicolumn{4}{c}{Based Model = Llama2-70B}\\
\cmidrule(lr){2-5} \cmidrule(lr){6-9} \cmidrule(lr){10-13}
& DQN & DMC & GPT3.5 & \textcolor{blue}{Agent} & DQN & DMC & GPT3.5 & \textcolor{blue}{Agent} & DQN & DMC & GPT3.5 & \textcolor{blue}{Agent}  \\
\midrule 
\textcolor{red}{Human}  &  -4.0   &  0.7  & -2.4   & \textcolor{red}{5.7}  & -4.0   & 0.7  & -2.4   & \textcolor{red}{5.7}   & -4.0   &  0.7  & -2.4   & \textcolor{red}{5.7} \\ \hline
\textcolor{blue}{Vanilla LLM}  &  -0.3   &  2.2  & -0.8   & \textcolor{blue}{-1.1}  & -2.2 &  1.7    & -0.9   & \textcolor{blue}{1.4}   & -0.8  & 3.4 & -0.4  & \textcolor{blue}{-2.2}  \\
\textcolor{blue}{Aggressive LLM}  &  -0.4   &  3.0  & -0.5   & \textcolor{blue}{-2.1}  & -2.0 &  2.8    & -1.0   & \textcolor{blue}{0.2}   & -1.6  & 7.6 & -1.2 & \textcolor{blue}{-4.8}\\
\textcolor{blue}{Conservative LLM} &  -0.7   &  2.9  & -0.9   & \textcolor{blue}{-1.3}  & -1.6 &  2.7    &-1.6   & \textcolor{blue}{0.5}   & -0.5  & 3.4 & -0.8 & \textcolor{blue}{-2.1}\\
\textcolor{blue}{Self-Consistency} &  -0.5   &  1.9  & -0.8   & \textcolor{blue}{-0.6}  & -2.8 &  2    & -0.7   & \textcolor{blue}{1.5}   & -1.0   & 3.8 & -0.9 &  \textcolor{blue}{-1.9} \\ 
\textcolor{blue}{ReAct}     &  -0.7    & 1.7     & -0.7    &  \textcolor{blue}{-0.3}       & -2.4  &  1.3     &  -1.1   & \textcolor{blue}{2.2}    &  -1.1 & 3.9 & -0.8 & \textcolor{blue}{-2.0} \\  
\textcolor{blue}{Reflexion}  &  -0.1   &  2.5 & -0.9   & \textcolor{blue}{-1.5}  & -2.6 &  2.1    & -0.7   & \textcolor{blue}{1.2}   & -1.2  & 4.7 & -0.9  & \textcolor{blue}{-2.6} \\ 
\textcolor{blue}{Multi-Agent} &  -1.1   &  2.3  & -0.3   & \textcolor{blue}{-0.9}  & -1.8 &  1.9    & -1.2   & \textcolor{blue}{1.1}   & -0.7  & 3.5 & -1.0 & \textcolor{blue}{-1.8}\\  
\textcolor{blue}{Agent-Pro} &  -1.5\tiny{$\downarrow $1.2}   &  1.4 \tiny{$\downarrow $0.8}  & -1.1 \tiny{$\downarrow $0.3}   & \textcolor{blue}{\textbf{1.2}} \tiny{$\uparrow$2.3}  & -3.9\tiny{$\downarrow $1.7} &  1.1 \tiny{$\downarrow $0.6}    & -1.5 \tiny{$\downarrow $0.6}  & \textcolor{blue}{\textbf{4.3}} \tiny{$\uparrow $2.9}   & -1.2 \tiny{$\downarrow $0.4}  & 3.1 \tiny{$\downarrow $0.3} & -0.5 \tiny{$\downarrow $0.1}   &  \textbf{\textcolor{blue}{-1.4}} \tiny{$\uparrow $0.8}  \\ 
- \textcolor{blue}{\texttt{w/o Learning}}  & -0.7  & 1.8   & -1.0   & \textcolor{blue}{-0.1}\tiny{$\uparrow $1}   & -3 &  1.5    & -1.2   & \textcolor{blue}{2.7}\tiny{$\uparrow $1.3}   & -0.3  & 3.3 &-1.2 & \textcolor{blue}{-1.8}\tiny{$\uparrow $0.4}\\ 

\bottomrule[0.5pt]
\end{tabular}
\caption{Each game contains four players. The first three are fixed as DQN, DMC, GPT-3.5, and the last one is the agent we need to evaluate: Agent-Pro or baselines. \texttt{Arrow} means comparison with Vanilla LLM.}\label{main results}

\end{table*}


\section{Game: Limit Texas Hold'em} \label{exp: Limit Texas Hold'em}
\textbf{Setups} In Limit Texas Hold'em, each player has two private cards and chooses from four actions: \texttt{Fold, Check, Call, Raise}. We set up matches among four players: DQN, DMC, GPT-3.5, and $\mathcal{X}$, where $\mathcal{X}$ represents the LLM-based agent we aim to evaluate, including Agent-Pro and baselines (\Cref{experiments details}). The prompts for baselines and Agent-Pro in~\Cref{baseline prompt for Texas Hold'em,agent-pro prompt for Texas Hold'em}. To enable Agent-Pro to learn within the game, we employ a total of 167 "training" game hands and 20 evaluation hands. Please refer~\Cref{setup for texas hold'em} for detail.

\textbf{Metrics} Similar to \Cref{policy optimization}, we sample 100 new game hands and allocate them to players. The players sequentially swap their hands and positions, generating 16 distinct permutations to eliminate the impact of chance and playing order. Lastly, we acquire 1600 games as the test set in total and calculate the average chip counts for four players. We provide detailed statistics in~\Cref{datset} regarding "training", evaluation, and test set.

\subsection{Results}
As shown in~\Cref{main results}, we report the final chip counts of various LLM-based agents against the other three players (DQN, DMC, GPT-3.5). The results indicate that Agent-Pro consistently outperforms RL-based agents e.g., DMC, and surpasses other LLM-based agents across numerous LLMs.

\textbf{Agent-Pro Surpasses LLM-based Agents and also Defeats RL-based Agents.} We observe that Agent-Pro achieves significant progress on GPT-3.5, GPT-4, and Llama2-70B, with an average score increase of +2 points. Besides, it surpasses specialized agents (DMC) on GPT-4, with an advantage of +3.2 points, and outperforms other LLM-based agents by a large margin (larger than 2.0 points). By analyzing the actions of Agent-Pro, we notice that it has learned to use multiple game techniques like humans. For instance, based on the analysis of the opponent's style in the \texttt{World Modeling}, it may coerce some cautious players into folding by bluffing or sometimes it may disguise itself to entice aggressive opponents to raise their bets.


\textbf{Belief Enhances Decision-making Capabilities in Dynamic Scenario.} Even without the learning process (policy-level reflection), Agent-Pro also can improve Vanilla LLM's performance by +0.9 points. For instance, on GPT-3.5 and GPT-4, it led to improvements of +1 points and +1.3 points, respectively, which already slightly surpasses most LLM-based agents. This improvement stems from the dynamic belief, which enables agents to promptly capture updates in community cards, changes in opponents' strategies, etc., thereby making more rational decisions. From the perspective of ReAct, our belief can also be seen as a dynamic thought process constructed based on the ToM framework, which endows agents with the ability to actively perceive internal and external beliefs and how they may change over time.

Besides, in~\Cref{ablation study}, we explore whether our evolution process could be replaced by few-shot learning, i,e., we add some demonstrations to the prompt of Vanilla LLM, and evaluate its results. We find that failed game trajectories can slightly improve its effectiveness, but not as significantly as our evolution strategy. In~\Cref{ablation study}, we also ablate the belief component from Agent-Pro but remain learning process. It shows that directly reflecting on the action sequence is quite unstable, and results in some vague and verbose behavioral instructions.


\begin{figure}[!t] 
    \centering
    \includegraphics[width=0.5\textwidth]{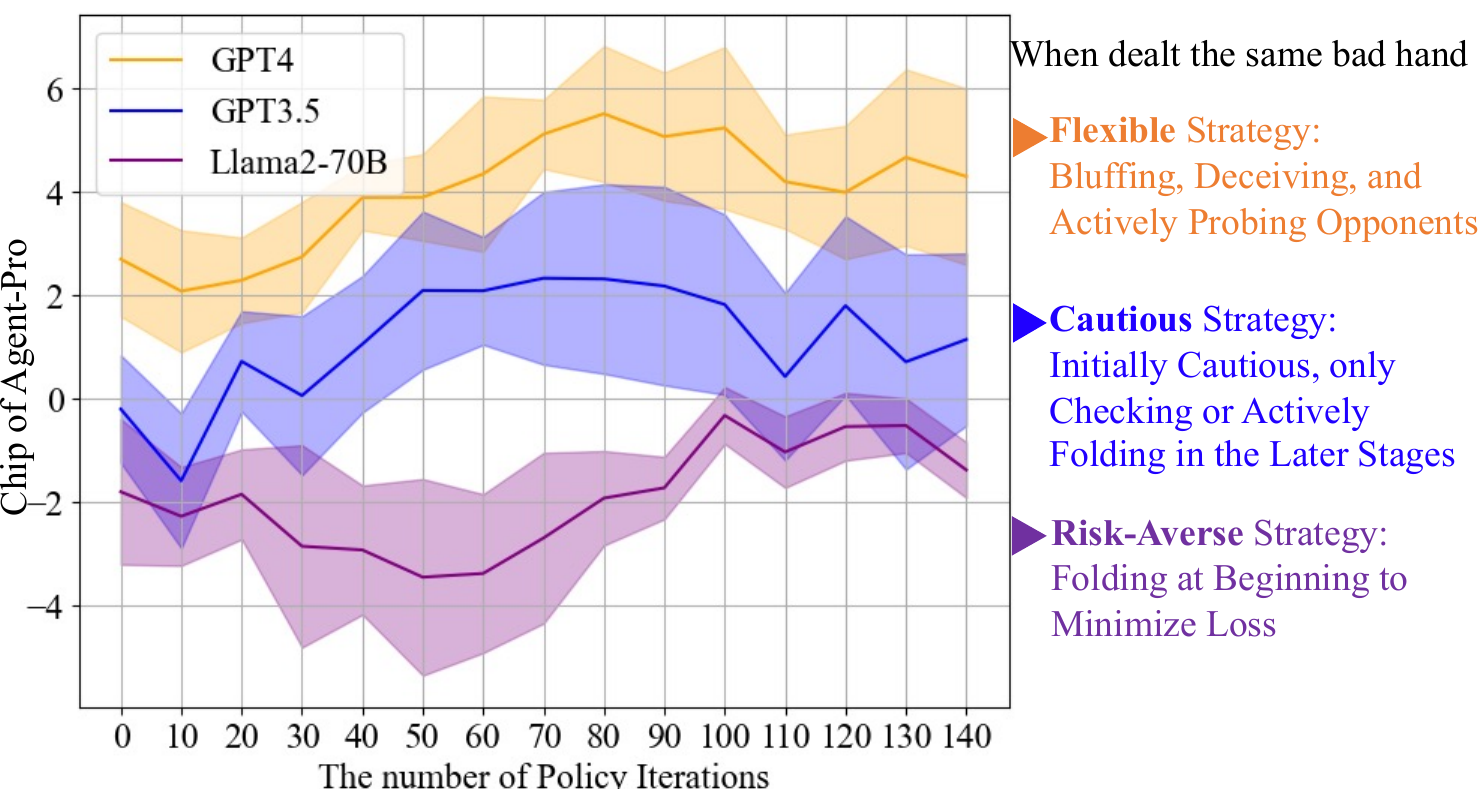}
    \caption{We report the relations between iteration number and the performance (average chips and its std).}
    \label{fig: learning trend}
\end{figure}

\begin{figure}[!t] 
    \centering
    \includegraphics[width=0.5\textwidth]{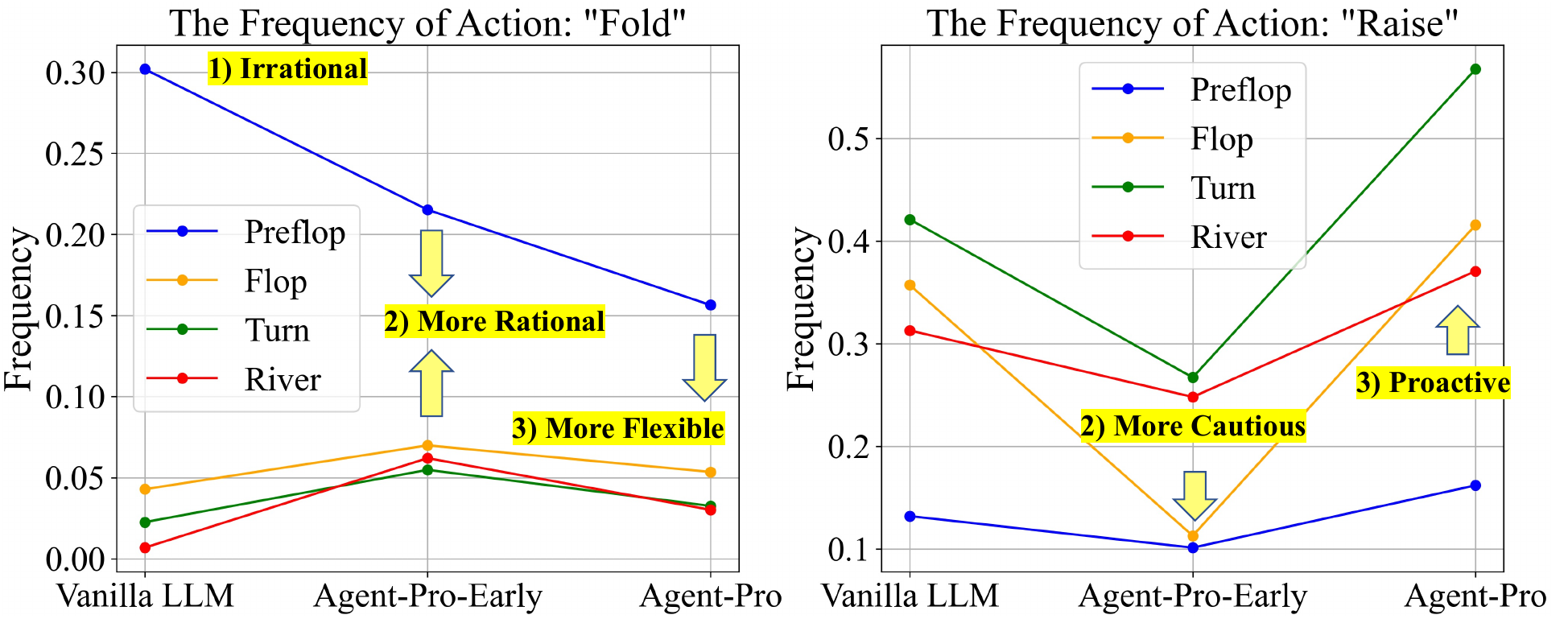}
    \caption{We analyze the \texttt{Fold} and \texttt{Raise} frequencies of three agents to illustrate the evolution of the strategy.}
    \label{fig: action}
\end{figure}

\subsection{Analysis on Learning Process}
We analyze the performance of Agent-Pro throughout the whole learning process. As shown in~\Cref{fig: learning trend}, Agent-Pro is evaluated every 10 iterations.

\textbf{Different LLM-based Agent-Pro Develops Diverse Strategies.} We observe that the learning curves of the three Agent-Pros exhibit significant differences. Agent-Pro based on GPT-4 and GPT-3.5 rapidly improves their performance in the early stages of learning, with a maximum increase of +2.1 and 2.3 chips respectively. In contrast, Llama-2-70B exhibits a dissimilar learning process, with performance initially declining in the first half and then improving (+0.6 chips) in the latter half. Analyzing the behaviors of the three agents, we discover that their strategic styles are entirely different. When dealt the same bad hand, the GPT-4-based Agent-Pro is relatively flexible and may bluff to probe opponents. GPT-3.5-based Agent-Pro tends to be cautious and may actively fold in the later stages, whereas the Llama-based Agent-Pro develops a highly conservative, risk-averse strategy. It concedes at the beginning of the game by opting to \texttt{Fold}, thereby losing only the initial few chips.



\subsection{Analysis on Policy Evolution}
We manually select 20 challenging games (Details in~\Cref{20cards}). Then, we test three agents on these 20 games: Agent-Pro in the early learning phase (Agent-Pro-Early), Agent-Pro, and Vanilla LLM. 

\textbf{How the Strategy Evolved.} We calculate the frequency of the most conservative action (\texttt{Fold}) and the most aggressive action (\texttt{Raise}) during the four stages of the game: \texttt{PreFlop, Flop, Turn, River}. As shown in~\Cref{fig: action}, we discuss how the strategy evolved. 1) The behavior of Vanilla LLM is rather \textbf{rigid}, \texttt{Folding} early in the game (\texttt{Preflop} stage) and ignoring subsequent community cards. 2) As learning progresses, Agent-Pro-Early becomes more \textbf{rational}, with a noticeable decrease in \texttt{Folding} frequency during the \texttt{Preflop} stage. It can observe the public cards in subsequent phases before deciding to \texttt{Folding}. Besides, Agent-Pro-Early is more \textbf{cautious}, with a significant decrease in the frequency of \texttt{Raising}. 3) After learning, Agent-Pro exhibits \textbf{flexible} and \textbf{proactive} behavior. Compared to Agent-Pro-Early, its \texttt{Fold} frequency in \texttt{Preflop} continues to decrease, but the frequency of \texttt{Raising} in all four stages has rebounded. This result demonstrates the evolution of the strategy: from irrational to rational, from conservative to flexible. A detailed case study is shown in~\Cref{case1 for Texas Hold'em,case2 for Texas Hold'em,case3 for Texas Hold'em,case4 for Texas Hold'em}.

\begin{figure}[!t] 
    \centering
    \includegraphics[width=0.46\textwidth]{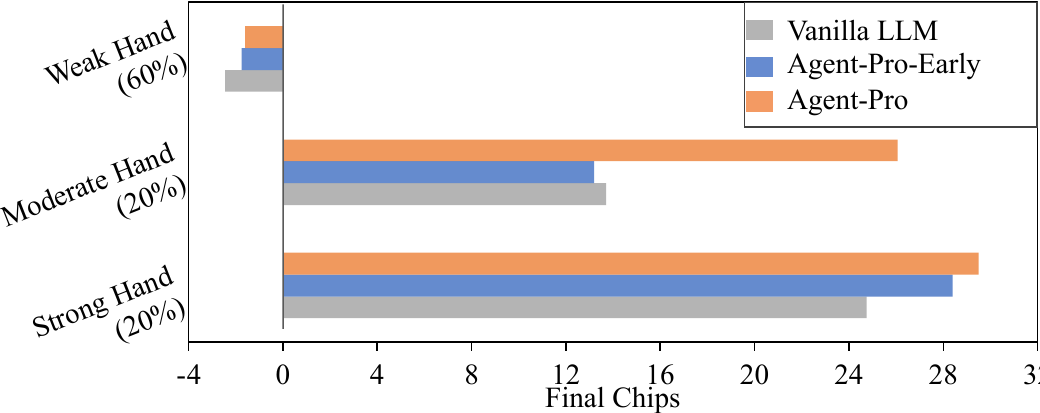}
    \caption{We categorize the agent's hands into three types: strong, medium-strength, and weak hands.}
    \label{goog-bad-card}
\end{figure}

\textbf{Win More, Lose Less.} As shown in~\Cref{goog-bad-card}, we categorize the hands dealt to the agent into three types: strong, medium, and weak hands, and record their performance separately. The results show that Agent-Pro can win more chips with strong hands and lose fewer chips with weak hands compared to Vanilla LLM. Notably, Agent-Pro significantly improves performance (> 80\%) with medium-strength hands,  which indicates that it learns advanced skills, expanding its capability boundaries.

\section{Discussions}
Beyond card games, Agent-Pro holds the potential to handle other complex tasks. Although we have only tested Agent-Pro in Blackjack and Texas Hold'em tasks, we believe there is great potential for it to be generalized to many complex tasks.
\begin{itemize}
    \item Firstly, the information set number for Texas Hold'em can reach up to $10^{162}$, far exceeding most tasks. Agent-Pro employs a belief-aware decision-making process to handle these challenges.
    \item Secondly, unlike most static environments, these strategic games are dynamic, with the game environment often changing in response to the opponent's actions. Agent-Pro establishes self-belief and world-belief to depict the dynamic environment.
    \item Lastly, due to the unobservable states of opponents, deceit and bluffing behaviors frequently occur in the game. In such a scenario, Agent-Pro manages to learn and evolve strategies, surpassing trained RL-based agents. Agent-pro adopts a policy-level reflection and optimization process to learn advanced game strategies.

\end{itemize}

\section{Conclusion}
We design an LLM-based agent, Agent-Pro, capable of learning and evolution in complex interactive tasks. It first constructs a dynamic belief for decision-making in uncertain scenarios. Then Agent-Pro reflects on its interactive experiences, corrects irrational beliefs, and summarizes its reflections into two instructions: behavioral guidelines and world descriptions for a new policy. Lastly, we evaluate Agent-Pro in two zero-sum games and observe that its decision-making capabilities significantly improve after learning from historical experiences.

\section*{Limitations}
Agent-Pro has presented a novel paradigm for designing an evolvable LLM-based agent, but we want to highlight that there remain some limitations or improvement spaces: 1) Dependency: the learning process of the Agent-Pro heavily relies on the capability of the foundational model, especially its reasoning and reflection abilities. In Texas Hold'em, the GPT-4-based Agent-Pro approaches the level of human players and surpasses DMC clearly, while GPT-3.5 and Llama2-70B-based Agent-Pro are still weaker than DMC, despite notable improvements. We plan to continue enhancing the capabilities of Agent-Pro based on weaker LLMs, aiming to achieve greater improvements even on smaller LLM models. 2) Performance: despite indispensable improvements, there may still be a significant gap between Agent-Pro and state-of-the-art algorithms (such as CFR plus) in gaming scenarios. In the future, we will continue to explore this issue and establish a set of benchmarks to evaluate their behaviors more comprehensively.

\section*{Acknowledgments}
This work was supported by the Nanjing Science and Technology Plan under Grants Y23002ZX01, the Key Research and Development Program of Zhejiang Province, China (No. 2024C01034).

\bibliography{custom}
\bibliographystyle{acl_natbib}

\appendix

\clearpage
\renewcommand\thefigure{\Alph{section}\arabic{figure}}    
\setcounter{figure}{0}    
\renewcommand\thetable{\Alph{section}\arabic{table}}    
\setcounter{table}{0}   

\noindent\textbf{Appendix}
\section{Experiment Details}\label{experiments details}

\subsection{LLMs} 
We employ the \texttt{GPT-3.5-Turbo-0613}, \texttt{GPT4-0613}, \texttt{Llama2-Chat-70B}~\citep{Touvron2023Llama2O} and \texttt{Qwen-72B}~\citep{bai2023qwen} to construct our agent. To make a fair comparison, we uniformly set the temperature to 1.0 for all experiments. For each test, we repeat it five times and report the average.


\subsection{Baselines} 
We compare Agent-Pro with many common LLM-based agent strategies, including Vanilla LLM, ReAct~\citep{Yao2022ReActSR}, Reflexion~\citep{Shinn2023ReflexionAA}, Self-Consistency~\citep{Wang2022SelfConsistencyIC}, Multi-agent Debate~\citep{Du2023ImprovingFA}.

\subsection{Setups For Blackjack} \label{setup for blackjack}
In Blackjack, players must decide to \texttt{hit} or \texttt{stand} based on their own hand, the dealer's face-up card. We simplify our approach by not incorporating \texttt{Verification} and \texttt{DFS-based Policy Evolution} since Blackjack is simple with a small state space. We collect 50 failed games for policy learning. We evaluate Agent-Pro and baselines on newly sampled 900 games. All prompts are presented in~\Cref{baseline prompt for blackjack,agent-pro prompt for blackjack}.

\subsection{Detailed Setup For Texas Hold'em}  \label{setup for texas hold'em}
The whole learning process is as follows: We first randomly allocate 500 game hands for 4 players and then select these failed game hands in which the agent loses a significant number of chips. We collect a total of 167 challenging game hands as the "training" set and 20 game hands as a development set for policy evaluation. Then Agent-Pro is instructed to conduct a learning process on these "training" instances, containing three phases: \textbf{Exploration-Reflection-Evolution}. 

Exploration ($\mathsection$~\ref{belief-aware process}): It randomly selects a game from "training" set to play with the latest policy and the belief-aware decision-making process. 

Reflection ($\mathsection$~\ref{policy-level reflection}): If Agent-Pro loses to its opponents, it immediately performs \texttt{Policy-Level Reflection} on this game and then updates to the new policy after passing \texttt{Verification}. 

Evolution ($\mathsection$~\ref{policy optimization}): We first sample 2 game hands from the development set to evaluate the new policy and calculate its $\Delta$ with $B$=8 for DFS. The process ends when the policy cannot be further improved, or all samples have been explored.

\section{Introduction of Two Games}\label{introduction of two games}
We selected the following two games as interactive environments.

\subsection{Blackjack} 
Blackjack~\footnote{\url{https://en.wikipedia.org/wiki/Blackjack}}
, also known as 21, is a popular card game that involves a dealer and a player. Players must decide whether to \texttt{hit} or \texttt{stand} based on their own hand, the dealer's face-up card, and the dealer's one hidden card. The objective is to beat the dealer without exceeding 21 points. For this game, we observe whether LLM-based agents can make rational decisions under uncertain scenarios.

\subsection{Limit Texas Hold'em} \label{Texas Hold'em description}
Limit Texas Hold'em is a popular card game~\footnote{\url{https://www.winamax.fr/en/poker-school_rules_limit-texas-hold--em}}
. The game commences with each player being dealt two private cards, which belong exclusively to the player and remain hidden from the others. Five community cards are then dealt face-up in a series of stages: a three-card \texttt{Flop}, followed by a single card on the \texttt{Turn} and another single card on the \texttt{River}. The player can choose from four actions: \texttt{Fold, Check, Call, Raise}. They aim to construct the best five-card poker hand possible using any combination of their private cards and community cards. 

\subsection{The Challenging of two Games}
Two games can evaluate the agent's capabilities from multiple dimensions:

\textbf{Handling Uncertainty in Environment:} Both games are imperfect information games and be used to assess the performance of LLM-based agents in face of uncertainty. For instance, in the game of Blackjack, the card hidden by the dealer introduces significant uncertainty. The agent needs to assess the risk and make decisions accordingly.

\textbf{Addressing Dynamic Environment:} Most real-world scenarios are dynamic rather than static. This requires the agent to capture environmental changes and adapt to them. For instance, in Texas Hold'em, in addition to one's own hand, the actions of opponents also greatly influence the agent's decisions. We evaluate whether Agent-Pro can handle such dynamic environments.

\textbf{Addressing Complexity:}  Blackjack is relatively simple, with an InfoSet number of 1000. In contrast, multi-player Limit Texas Hold'em is very complex, with its Infoset number reaching up to $10^{14}$~\citep{zha2019rlcard}. We analyze Agent-Pro's learning capacity in such intricate scenarios.

\begin{table}[t!]\small
    \centering
    \setlength\tabcolsep{3pt} 
    \begin{tabular}{ l    l l l }
    \toprule[1pt]
    \textbf{\makecell[l]{Game}}  & \textbf{\makecell[c]{\#Train}} &   \textbf{\#Dev} &\textbf{\#Test}\\ \hline
    \makecell[l]{Blackjack}     & 50   & -  & 900 \\
    \makecell[l]{Texas Hold’em} & 167  & 20 & 1600\\
    \bottomrule[1pt]
    \end{tabular}
    \caption{The sample sizes of the Training, Development, and Testing sets for the two games, where the training set is utilized for exploration and reflection, the development set for policy evaluation, and the test set for assessing the effectiveness of all methods.}\label{datset}
\end{table}

\begin{table}[t!]\small
    \centering
    \setlength\tabcolsep{2pt} 
    \begin{tabular}{ l | l l l l }
    \toprule[1pt]
     \makecell[c]{\textbf{Strategy}} & DQN  & DMC & GPT-3.5 & \textcolor{blue}{Agent}\\ \hline
    \textcolor{black}{Vanilla LLM}   & -2.2  & 1.7 & -0.9 & \textcolor{blue}{1.4}   \\
    \textcolor{black}{- \texttt{w/ 3 win shots}}     & -2.4  & 2.6 & -1.1 & \textcolor{red}{0.9} \tiny{$\downarrow$0.5}  \\ 
    \textcolor{black}{- \texttt{w/ 3 lose shots}}     & -2.6  & 1.8 & -1.2 & \textcolor{blue}{2.0} \tiny{$\uparrow$+0.6}  \\ 
    \textcolor{black}{- \texttt{w/ 3 win+3 lose shots}}     & -1.9  & 1.9 & -1.5 & \textcolor{blue}{1.5} \tiny{$\uparrow$+0.1}  \\ \hline
    \textcolor{black}{Agent-Pro}     & -3.9  & 1.1 & -1.5 & \textcolor{blue}{4.3} \tiny{$\uparrow$2.9}  \\
    \textcolor{black}{- \texttt{w/o Belief}}     & -3.3  & 1.5 & -0.7 & \textcolor{blue}{2.5} \tiny{1.1}  \\ 
    \textcolor{black}{- \texttt{w/o Learning}}  & -3 &  1.5    & -1.2   & \textcolor{blue}{2.7}\tiny{$\uparrow $1.3}  \\
    \bottomrule[1pt]
    \end{tabular}
    \caption{Up: We compare the performance of Agent-Pro with Vanilla LLM with few-shot demonstrations. Each demonstration contains a complete trajectory and final results. Down: We ablate the dynamic belief or learning process from Agent-Pro and evaluate its results.}\label{ablation study}
\end{table} 

\section{Complementary Experiments}
To better investigate the performance of Agent-Pro, we design some ablation experiments.
\subsection{Whether Few-shot Learning Can Handle Such Complex Interaction}
 First, we compare Agent-Pro with Few-shot Agent: we randomly select some winning and losing game trajectories and their final results as demonstrations in the prompt. Then we evaluate them on the test set. As shown in~\Cref{ablation study}, we observe that winning trajectories seem to have no effect (-0.5), while losing trajectories can slightly improve the final performance (+0.6). This phenomenon is quite intriguing, indicating that these winning demonstrations do not seem to enhance the agent's decision-making ability in such complex scenarios. This may be because these winning demonstrations are relatively simple for the vanilla agent, which is already capable of winning these games, so including them in the prompt does not provide it with any additional insights. In contrast, those failing trajectories instead promote agents to reflect and adjust their behaviors, improving the final results.

\subsection{Policy-Level Reflection without Belief}
We ablate the dynamic belief module, i.e., conducting policy-level reflection directly on the action sequences, state sequences, and final results. Then it also summarizes prompt instructions for policy updates. As shown in~\Cref{ablation study}, we observe that after removing, Agent-Pro's chips drop from 4.3 to 2.5, although there is still an improvement of +1.1 compared to Vanilla LLM. Upon closely examining the \texttt{Behavioral Guideline} and \texttt{World Modeling} it generated, we observe that compared to Agent-Pro, its content is rather vague and verbose, lacking in specificity and conciseness. These results indicate that dynamic belief can enhance both decision-making and policy-level reflection capabilities. Below, we provide two similar instructions, one from Agent-Pro and the other from Agent-Pro without Belief.

\begin{lstlisting}
(*@\textbf{A Learned Instruction From Agent-Pro}@*)
When holding a weak hand, adopting a conservative approach and waiting for the flop can be wise. This strategy allows for the possibility of the community cards improving your hand. However, if the flop doesn't enhance your hand's strength, folding to minimize losses becomes the prudent choice. During this period, maintaining a low profile and avoiding aggressive actions like raising is advisable.

(*@\textbf{A Similar Instruction From Agent-Pro w/o Belief}@*)
In situations where the strength of one's hand isn't exactly what one might call robust or particularly promising, it could potentially be somewhat beneficial, or at least not entirely disadvantageous, to entertain the notion of adopting a stance that leans more towards the side of caution.... the unveiling of the community cards.....
\end{lstlisting}


\begin{table}[t!]\small
    \centering
    \setlength\tabcolsep{4pt} 

    \begin{tabular}{ l c c  c}
    \toprule[1pt]
    \textbf{Strategy} & \textbf{DQN} & \textbf{DMC}  &\textbf{Human}  \\ 
    Win-rate $\uparrow$(\%)       &  \textcolor{black}{40.0}   & \textcolor{black}{41.9}   & \textcolor{black}{37.1}   \\
    \bottomrule[1pt]
    \end{tabular}
    \caption{We evaluate two RL-based agents, and the human player on the same 900 games.}\label{other result of blackjack}
\end{table}

\subsection{Detailed Analysis Experiments}
As shown in~\Cref{20cards}, we manually select 20 challenging sets of hands, each with a significant difference in the hands of four players, and then assess the performance of Agent-Pro and Vanilla LLM. 
\begin{table*}[t!]\small
    \centering
    \begin{tabular}{l|ll|ll|ll|ll|lll|l|l}
        \toprule[1pt]
        \multirow{3}*{\thead{Hand\\Strength}} & \multicolumn{8}{c|}{Hand} & \multicolumn{5}{c}{Community Cards}\\
        \cmidrule(lr){2-9}\cmidrule(lr){10-14}
        & \multicolumn{2}{c|}{DQN} & \multicolumn{2}{c|}{DMC} & \multicolumn{2}{c|}{GPT3.5} & \multicolumn{2}{c|}{\textcolor{blue}{Agent}} & \multicolumn{3}{c|}{Flop} & Turn & River\\
        \midrule
        \multirow{4}*{Strong} & H5 & S4 & D6 & DQ & S9 & C2 & \textcolor{blue}{CA} & \textcolor{blue}{HA} & H7 & CQ & CT & H3 & S3  \\
         & DK & S5 & HK & D2 & S7 & H4 & \textcolor{blue}{DA} & \textcolor{blue}{HA} & DQ & D9 & DT & C6 & D7  \\
         & D3 & C8 & HA & HT & H5 & S9 & \textcolor{blue}{DQ} & \textcolor{blue}{DJ} & D4 & CK & H7 & CQ & C5  \\
         & SA & H9 & C6 & S8 & S3 & SJ & \textcolor{blue}{HT} & \textcolor{blue}{CK} & D7 & C5 & C4 & C3 & D2  \\
        \cmidrule(lr){2-14}
        \multirow{4}*{Moderate} & HJ & CQ & S7 & SA & C3 & D5 & \textcolor{blue}{H3} & \textcolor{blue}{CA} & C2 & H9 & S3 & D9 & C8  \\
         & H5 & C5 & DJ & H9 & S6 & D2 & \textcolor{blue}{HK} & \textcolor{blue}{H2} & DA & DK & SK & C3 & H8  \\
         & H5 & D6 & DT & CT & C9 & C4 & \textcolor{blue}{S9} & \textcolor{blue}{C8} & S6 & S7 & HT & HQ & HJ  \\
         & C3 & D5 & H3 & CA & HJ & CQ & \textcolor{blue}{S7} & \textcolor{blue}{SA} & C2 & H9 & S3 & D9 & C8  \\
        \cmidrule(lr){2-14}
        \multirow{12}*{Weak} & S3 & SJ & HT & CK & SA & H9 & \textcolor{blue}{C6} & \textcolor{blue}{S8} & D7 & C5 & C4 & C3 & D2  \\
         & S3 & C8 & H7 & S2 & DA & CA & \textcolor{blue}{D7} & \textcolor{blue}{D5} & H8 & HJ & SJ & ST & D9  \\
         & DK & D5 & SJ & C6 & D9 & S3 & \textcolor{blue}{H2} & \textcolor{blue}{C8} & CA & CQ & CT & D8 & C2  \\
         & H5 & H8 & HA & S9 & C6 & D9 & \textcolor{blue}{C5} & \textcolor{blue}{H3} & D3 & C3 & HQ & S3 & DA  \\
         & HA & S9 & C6 & D9 & C5 & H3 & \textcolor{blue}{H5} & \textcolor{blue}{H8} & D3 & C3 & HQ & S3 & DA  \\
         & H2 & C8 & DK & D5 & SJ & C6 & \textcolor{blue}{D9} & \textcolor{blue}{S3} & CA & CQ & CT & D8 & C2  \\
         & H7 & S2 & DA & CA & D7 & D5 & \textcolor{blue}{S3} & \textcolor{blue}{C8} & H8 & HJ & SJ & ST & D9  \\
         & HA & HT & H5 & S9 & DQ & DJ & \textcolor{blue}{D3} & \textcolor{blue}{C8} & D4 & CK & H7 & CQ & C5  \\
         & DA & HA & DK & S5 & HK & D2 & \textcolor{blue}{S7} & \textcolor{blue}{H4} & DQ & D9 & DT & C6 & D7  \\
         & DT & CT & C9 & C4 & S9 & C8 & \textcolor{blue}{H5} & \textcolor{blue}{D6} & S6 & S7 & HT & HQ & HJ  \\
         & D6 & DQ & S9 & C2 & CA & HA & \textcolor{blue}{H5} & \textcolor{blue}{S4} & H7 & CQ & CT & H3 & S3  \\
         & HK & H2 & H5 & C5 & DJ & H9 & \textcolor{blue}{S6} & \textcolor{blue}{D2} & DA & DK & SK & C3 & H8  \\
        \bottomrule[1pt]
    \end{tabular}
    \caption{Each card consists of a letter representing the suit ("S", "H", "D", "C") and a number representing size ("2", "3", "4", "5", "6", "7", "8", "9", "T", "J", "Q", "K", "A"). Among them, "S" represents Spade, "H" represents Heart, "D" represents Diamond, "C" represents Club, and "T" represents 10. }
    \label{20cards}
\end{table*}

\section{Related Works}
\subsection{LLM-based Application}
Large language models (LLMs), pre-trained on extensive corpora, have demonstrated robust language comprehension and reasoning capabilities. Benefiting from this, researchers have designed a plethora of agent systems built upon LLM, achieving promising results~\citep{xi2023rise}. \citet{schick2023toolformer, wu2023visual, shen2023hugginggpt,wu2023precedent,wu2024biased} have harnessed the planning capabilities of LLMs to invoke specialized models and tools for task-solving. Some open-source projects, e.g., AutoGPT\footnote{\url{https://github.com/Significant-Gravitas/Auto-GPT}}, gentGPT\footnote{\url{https://github.com/reworkd/AgentGPT}}, BabyAGI\footnote{\url{https://github.com/yoheinakajima/babyagi}}, BMTools\footnote{\url{https://github.com/OpenBMB/BMTools}}, ChatArena\footnote{\url{https://github.com/chatarena/chatarena}}, LangChain\footnote{\url{https://github.com/hwchase17/langchain}} have developed an LLM-based assistant. Further, \citep{qin2023tool, qin2023toolllm, shen2023taskbench} have empowered LLM to autonomously invoke the APIs for daily life scenarios. Besides, leveraging the code generation capabilities of LLMs, researchers have designed multi-agent collaborative systems~\citep{li2023camel, Chen2023AgentVerseFM, Hong2023MetaGPTMP, Wu2023AutoGenEN} for complex tasks, such as software development. Unlike these task-specific agents that require manually specified behavior protocols, our agents can understand tasks through interaction with the environment. It can optimize its behavioral strategy from past experiences, accomplishing the task more effectively.

\subsection{LLMs For Interactive Scenarios}
Beyond these applications, LLMs have also been utilized in interactive settings~\citep{durante2024interactive}. ReAct~\citep{Yao2022ReActSR} integrates reasoning, action, and observation into the problem-solving process. \citet{Park2023GenerativeAgents} introduces generative agents that can simulate human behavior. \citet{Fu2023ImprovingLM} show LLMs can improve each other in a negotiation scenario. \citet{zhao2023expel,chen2023introspective} propose an experiential learner gathering experiences and extracting from a collection of training tasks. \citet{fan2023can} explored the capability of LLMs to make rational decisions in game-theoretic scenarios. Besides, some studies have designed sophisticated LLM-based agents for large-scale games, including StarCraft~\citep{ma2023large}, Minecraft~\citep{wang2023voyager,gong2023mindagent}, Leduc Hold'em~\citep{guo2023suspicion}, strategy-based gaming~\citep{Bakhtin2022HumanlevelPI,xu2023exploring,wang2023avalon,xu2023language, lore2023strategic} and application for legal contexts~\citep{wu2020biased, wu2022towards}.

\subsection{Improving the Quality of LLM Responses}
Enhancing the quality of responses from LLMs has garnered significant attention within the community. We categorize the strategies into two methodologies: 1. Developing superior reasoning architectures. First, Chain-of-Thoughts~\citep{wei2022chain} elicits LLM's reasoning ability. Works as Least-to-Most~\citep{Zhou2022LeasttoMostPE}, Tree of Thoughts~\citep{Yao2023TreeOT}, Graph of Thoughts~\citep{besta2023graph} have explored diverse problem-solving procedures and reasoning architectures, significantly enhancing the performance of LLM-based agents. 2. Refining the output of LLMs. Researchers have proposed post-hoc prompting strategies to iteratively refine the outputs of LLMs~\citep{Pan2023AutomaticallyCL}, including Reflexion~\citep{Shinn2023ReflexionAA}, Self-Refine~\citep{Madaan2023SelfRefineIR, Paul2023REFINERRF, Huang2022LargeLM}, Self-Contrast~\citep{zhang2024self}, etc. However, these self-correction strategies are performed at the action-level, whereas our agent operates at the policy-level, making it more suited for interactive environments. 

Additionally, Supervised Fine-Tuning (SFT) offers another avenue for enhancing LLMs, though it depends on human-annotated data. Recently, \citet{chen2024self} introduced Self-Play fIne-tuNing (SPIN), a novel fine-tuning approach. SPIN's foundation is a self-play mechanism that enables the LLM to refine its abilities by engaging with its own variations.

\subsection{Automatic Prompt Optimization}
In addition to optimizing the outputs of LLMs, many researchers also enhance the performance of LLMs by searching for a more effective prompt~\citep{zhou2022large, hsieh2023automatic, guo2023connecting, wang2023promptagent}. APO~\citep{pryzant2023automatic} emulates the process of gradient optimization. It calculates the "gradients" of the current prompt by analyzing the instances that are inaccurately predicted by this prompt. Furthermore, \citet{yang2023large} and \citet{ye2023prompt} evaluate each candidate prompt using the training set and iteratively optimize the prompts based on the evaluation results. \citet{cheng2023black} train a Sequence-to-Sequence model to translate an imperfect prompt into a better one. \citet{brooks2023large} and \citet{zhang2023large} combine reinforcement learning with prompt updating, demonstrating promising results. We extend these prompt optimization techniques to more complex interactive gaming environments, learning a robust behavioral strategy through policy-level reflection and search. Furthermore, our agent must constantly consider changes in the environment and the styles of opponents, thereby dynamically adjusting the content of the prompts.

\subsection{Benchmarking LLM-based Agent}
To develop an effective evaluation method for LLMs and their capabilities as agents, numerous researchers have concentrated on establishing benchmarks. SmartPlay~\citep{wu2024smartplay} introduces a benchmark from 6 diverse games with language descriptors for visual observation.  Clembench ~\citep{chalamalasetti-etal-2023-clembench} employs Dialogue Games as testing tools, enabling rapid evaluations across a broad of models. Furthermore,  \citet{liu2024agentbench} unveil AGENTBENCH, a comprehensive benchmark that outlines eight distinct environments to assess LLMs. These benchmarks play a crucial role in evaluating both LLMs and LLM-based agents. In our future work, we aim to further evaluate Agent-Pro utilizing these benchmarks.

\section{Detailed Prompts}
We provide detailed prompt designs for two games, including baselines and Agent-Pro in~\Cref{baseline prompt for blackjack}, \ref{agent-pro prompt for blackjack}, \ref{baseline prompt for Texas Hold'em} and \ref{agent-pro prompt for Texas Hold'em}.
\begin{figure*} 
\subsection{Baseline's Prompts For Blackjack}\label{baseline prompt for blackjack}
\begin{lstlisting}
(*@\textbf{Game Rule: }@*) 
### Game Rules
1. Please try to get your card total to as close to 21 as possible, without going over, and still having a higher total than the dealer.
2. If anyone's point total exceeds 21, he or she loses the game. 
3. You can only choose one of the following two actions: {"Stand", "Hit"}. If you choose to Stand, you will stop taking cards and wait for the dealer to finish. If you choose to Hit, you can continue to take a card, but there is also the risk of losing the game over 21 points. 
4. After all players have completed their hands, the dealer reveals their hidden card. Dealers must hit until their cards total 17 or higher.

(*@\textbf{Game Information: }@*) The dealer's face-up card is {(*@\color{blue}{Dealer-Card}@*)}. The dealer has another hidden card. You don't know what it is. Your current cards are {(*@\color{blue}{Player-Card}@*)}.

---------- (*@\textbf{Prompt For Vanilla LLM}@*) --------------
You are a player in blackjack. Please beat the dealer and win the game.
### {(*@\color{blue}{Game Rules}@*)} 
### {(*@\color{blue}{Game Information}@*)} 
### Please output your action in following format: ###My action is {Your action}, without any other text.

------------- (*@\textbf{Prompt For Radical LLM}@*) --------------
You are an aggressive player of blackjack who likes to take risks to earn high returns. Please beat the dealer and win the game.
### {(*@\color{blue}{Game Rules}@*)}
### {(*@\color{blue}{Game Information}@*)} 
### Please output your action in following format: ###My action is {Your action}, without any other text.

------------- (*@\textbf{Prompt For ReAct}@*) --------------
You are a player in blackjack. Please beat the dealer and win the game.
### {(*@\color{blue}{Game Rules}@*)}
### {(*@\color{blue}{Game Information}@*)} 
### Please first think and reason about the current hand and then generate your action as follows: ###My thought is {Your Thought}. My action is {Your action}.

------------- (*@\textbf{Prompt For Reflexion}@*) --------------
You are a player in blackjack. Please beat the dealer and win the game.
### {(*@\color{blue}{Game Rules}@*)}
### {(*@\color{blue}{Game Information}@*)} 
### Please first think about the current hand and then generate your action in following format: ###My thought is {Your thought}. My action is {Your action}.
Assistant: {(*@\color{blue}{LLM Response}@*)}. My action is {(*@\color{blue}{LLM Response}@*)}
### Please carefully check the response you just output, and then refine your answer. The final output is also in following format: ###My thought is {Your thought}. My action is {Your action}.
\end{lstlisting}
\end{figure*}

\begin{figure*} 
\subsection{Agent-Pro's Prompt For Blackjack}\label{agent-pro prompt for blackjack}
\begin{lstlisting}
(*@\textbf{Game Rule: }@*) 
### Game Rules
1. Please try to get your card total to as close to 21 as possible, without going over, and still having a higher total than the dealer.
2. If anyone's point total exceeds 21, he or she loses the game. 
3. You can only choose one of the following two actions: {"Stand", "Hit"}. If you choose to Stand, you will stop taking cards and wait for the dealer to finish. If you choose to Hit, you can continue to take a card, but there is also the risk of losing the game over 21 points. 
4. After all players have completed their hands, the dealer reveals their hidden card. Dealers must hit until their cards total 17 or higher.

(*@\textbf{Game Information: }@*) The dealer's face-up card is {(*@\color{blue}{Dealer-Card}@*)}. The dealer has another hidden card. You don't know what it is. Your current cards are {(*@\color{blue}{Player-Card}@*)}.

------------- (*@\textbf{Prompt For Agent-Pro}@*) --------
You are a player in blackjack. Please beat the dealer and win the game.
### {(*@\color{blue}{Game Rules}@*)}
### {(*@\color{blue}{Game Information}@*)} 
### {(*@\color{blue}{Behavioral Guideline: {Goal}, {Strategy}, {Demonstration}@*)}
### {(*@\color{blue}{World Modeling: {Rule Description}@*)}
### Please read the behavoiral guideline and world modeling carefully. Then you should analyze your own cards and your strategies in Self-belief and then analyze the dealer cards in World-belief. Lastly, please select your action from {"Stand", "Hit"}. 
### Output Format: Self-Belief is {Belief about youself}. World-Belief is {Belief about the dealer}. My action is {Your action}. Please output in the given format.

------------- (*@\textbf{Prompt For Policy-Level Reflection}@*) --------------
### {(*@\color{blue}{Game Rules}@*)}
### Game Record: {(*@\color{blue}{Game Record, Belief Sequences, Final Result}@*)}
### You are a seasoned blackjack expert, and you need to carefully reflect on the following record of this losing game:

(*@\textbf{Correctness}@*): Whether its beliefs about yourself, the game, and the dealer align with the final results.
(*@\textbf{Consistency}@*): Whether each belief and action is self-contradictory.
(*@\textbf{Reasons}@*): Reflect on why you lost to your dealer, which beliefs and actions are problematic, and what the underlying reasons are.

### Output Format: I analyze this game as follows: {Your analysis about the game and belief}.

------------- (*@\textbf{Prompt For Generating Behavioral Guideline and World Modeling}@*) --------------
### Game Record: {(*@\color{blue}{Game Record, Belief Sequences, Final Result}@*)}
### Policy-Level Reflection: {(*@\color{blue}{Reflection}@*)}
Following the previous rigorous analysis, you should distill and articulate a set of Behavioral Guidelines and World Modeling. The Behavioral Guideline is about what you consider to be a more reasonable and effective behavioral strategy and suggestions. World Modeling is about the description of the game and the dealer. 

Here are some suggestions for you: 

(*@\textbf{Behavioral Guideline}@*)
1-(*@\color{blue}{Goal}@*): Please summarize the detailed goal based on your reflection ... 
2-(*@\color{blue}{Strategy}@*): What kind of strategy can lead you to win in similar games ... 
3-(*@\color{blue}{Demonstration}@*): Can this game be considered a typical example to be preserved for future reference ... 
(*@\textbf{World Modeling}@*)
1-(*@\color{blue}{Rule-Description}@*): Based on the recent reflection, describe any game rules or details that are easy to overlook ... 
\end{lstlisting}
\end{figure*}

\begin{figure*} 
\subsection{Baseline's Prompts For Limited Texas Hold'em} \label{baseline prompt for Texas Hold'em}
\begin{lstlisting}
(*@\textbf{Game Settings: }@*) 
1- You are playing the Limit Texas poker game. In this game, there are 4 players from 0 to 3, and your role is player 3. 
2- The number of chips every player has is infinite.  
3- You just need to win new chips in the competition as much as possible.
4- The actions you can choose are ['call', 'raise', 'fold', 'check']

(*@\textbf{Game Information: }@*) 
Your current hands are {(*@\color{blue}{Private Cards}@*)}. 
The current stage: {(*@\color{blue}{Stage}@*)}. Public cards are {(*@\color{blue}{Public Cards}@*)}. 
Number of chips all players have invested are {(*@\color{blue}{Inveseted Chip List}@*)}. 
Available actions you can choose are {(*@\color{blue}{Available Actions}@*)}. 
Previous actions of all players are: {(*@\color{blue}{Preflop: Actions Sequences, Flop: Actions ...}@*)}. 

---------- (*@\textbf{Prompt For Vanilla LLM}@*) --------------
You are a player in Limited Texas Hold'em. Beat your opponents and win the game.
### {(*@\color{blue}{Game Rules}@*)} 
### {(*@\color{blue}{Game Information}@*)} 
### Output your action in following format: {"action": " "} without any other text.

---------- (*@\textbf{Prompt For Aggressive LLM}@*) --------------
You are an aggressive player of limited Texas Hold'em who likes to take risks to earn high returns. Please beat your opponents and win the game.
### {(*@\color{blue}{Game Rules}@*)} 
### {(*@\color{blue}{Game Information}@*)} 
### Output your action in following format: {"action": " "} without any other text.

---------- (*@\textbf{Prompt For Conservative LLM}@*) --------------
You are a conservative player of limited Texas Hold'em who is risk averse and prefers more certainty. Please beat your opponents and win the game.
### {(*@\color{blue}{Game Rules}@*)} 
### {(*@\color{blue}{Game Information}@*)} 
### Output your action in following format: {"action": " "} without any other text.

---------- (*@\textbf{Prompt For ReAct}@*) --------------
You are a player in Limited Texas Hold'em. Beat your opponents and win the game.
### {(*@\color{blue}{Game Rules}@*)} 
### {(*@\color{blue}{Game Information}@*)} 
### Please first think and reason about the current state and then generate your action as follows: ###My thought is {Your Thought}, and my action is {"action": " "}

---------- (*@\textbf{Prompt For Reflection}@*) --------------
You are a player in Limited Texas Hold'em. Beat your opponents and win the game.
### {(*@\color{blue}{Game Rules}@*)}
### {(*@\color{blue}{Game Information}@*)} 
### Please first think and reason about the current state and then generate your action as follows: ###My thought is {Your Thought}, and my action is {"action": " "}.

Assistant: {(*@\color{blue}{LLM Response}@*)}.

### Please carefully check the thought and the action you just output, and then refine your answer. The final output is also in the same format: ###My revised thought is {Your Thought}. My revised action is {"action": " "}.
\end{lstlisting}
\end{figure*}

\begin{figure*} 
\subsection{Agent-Pro's Prompt For Limited Texas Hold'em} \label{agent-pro prompt for Texas Hold'em}
\begin{lstlisting}
(*@\textbf{Game Settings: }@*) 
1- You are playing the Limit Texas poker game. In this game, there are 4 players from 0 to 3, and your role is player 3. 
2- The number of chips every player has is infinite.  
3- You just need to win new chips in the competition as much as possible.
4- The actions you can choose are ['call', 'raise', 'fold', 'check']

(*@\textbf{Game Information: }@*) 
Your current hands are {(*@\color{blue}{Private Cards}@*)}. 
The current stage: {(*@\color{blue}{Stage}@*)}. Public cards are {(*@\color{blue}{Public Cards}@*)}. 
Number of chips all players have invested are {(*@\color{blue}{Inveseted Chip List}@*)}. 
Available actions you can choose are {(*@\color{blue}{Available Actions}@*)}. 
Previous actions of all players are: {(*@\color{blue}{Preflop: Actions Sequences, Flop: Actions ...}@*)}. 

------------- (*@\textbf{Prompt For Agent-Pro}@*) --------
You are a player in Limited Texas Hold'em. Beat your opponents and win the game.
### {(*@\color{blue}{Game Rules}@*)}
### {(*@\color{blue}{Game Information}@*)} 
### {(*@\color{blue}{Behavioral Guideline: Goal, Strategy, Demonstration}@*)}
### {(*@\color{blue}{World Modeling: Rule Description, Opponents Description}@*)}
### Please read the behavoiral guideline and world modeling carefully. Then, following their instructions, you should analyze your own cards and your strategies in Self-belief and then analyze and reason about your opponents in World-belief. Lastly, output your action.
### Output Format: Self-Belief is {belief about youself}. World-Belief is {belief about the all opponents}. My action is {"action": " "}. Please output in the given format.

------------- (*@\textbf{Prompt For Policy-Level Reflection}@*) --------------
### {(*@\color{blue}{Game Rules}@*)}
### {(*@\color{blue}{Game Information}@*)} 
### Game Record: {(*@\color{blue}{Game Record, Belief Sequences, Final Result}@*)}
### You are a seasoned Limited Texas Hold'em expert, and you need to carefully reflect on the following record of this losing game:

(*@\textbf{Correctness}@*): Whether its beliefs about yourself, the game, and the opponents align with the final results.
(*@\textbf{Consistency}@*): Whether each belief and action is self-contradictory.
(*@\textbf{Rationality}@*): Whether the beliefs accurately reflect the underlying intentions behind the opponents. 
(*@\textbf{Reasons}@*): Reflect on why you lost to your opponents, which beliefs and actions are problematic, and what the underlying reasons are.

### Output Format: I analyze this game as follows: {Your analysis about the game and belief}.

------------- (*@\textbf{Prompt For Generating Behavioral Guideline and World Modeling}@*) --------------
### Game Record: {(*@\color{blue}{Game Record, Belief Sequences, Final Result}@*)}
### Policy-Level Reflection: {(*@\color{blue}{Reflection}@*)}
Following the previous rigorous analysis, you should distill and articulate a set of Behavioral Guidelines and World Modeling. The Behavioral Guideline is about what you consider to be a more reasonable and effective behavioral strategy and suggestions. World Modeling is about the description of the game and all opponents. 

Here are some suggestions for you: 

(*@\textbf{Behavioral Guideline}@*)
1-(*@\color{blue}{Goal}@*): Please summarize the detailed goal based on your reflection ... 
2-(*@\color{blue}{Strategy}@*): What kind of strategy can lead you to win in similar games ... 
3-(*@\color{blue}{Demonstration}@*): Can this game be considered a typical example to be preserved for future reference ... 
(*@\textbf{World Modeling}@*)
1-(*@\color{blue}{Rule-Description}@*): Based on the recent reflection, describe any game rules or details that are easy to overlook ... 
2-(*@\color{blue}{Opponent-Modeling}@*): Based on each opponent's action and the real hands shown at the end of the game, what do you think their style and strategy are ... 
\end{lstlisting}
\end{figure*}

\section{Case Study}
As shown in~\Cref{Blackjack Case1}, \ref{Blackjack Case2}, \ref{Blackjack Case3}, \ref{Blackjack Case4} we provide four cases for Blackjack. We visualize the difference in their solving steps between Agent-Pro and ReAct when using Qwen-72B. Besides, we also provide four cases for Limited Texas Hold'em in~\Cref{case1 for Texas Hold'em}. These cases demonstrate that Agent-Pro, after learning, has significantly improved in understanding task rules, mastering techniques, and dealing with uncertain environments.
\begin{figure*}[!ht]
\subsection{Cases For Blackjack}
    \centering
    \setlength\tabcolsep{4pt} 
     Agent-Pro is Better at Handling the Uncertainties of the Environment.
     \small
    \begin{tabular}{l| p{6cm} | p{7cm} }
    \toprule[1pt]
    \textbf{State} & \makecell[c]{\textbf{ReAct Vs. Dealer }} & \makecell[c]{\textbf{Agent-Pro Vs. Dealer}}  \\ \hline
    \textcolor{blue}{Initial Card} & \multicolumn{2}{p{13cm}}{The dealer's face-up card is \textbf{Spade 4}.  \  \  \  \  Your current cards are \textbf{Heart 5, Club J}.} \\ \hline
    \textcolor{blue}{Thought} 
    & My thought is I currently have a total of 15 points with my two cards, which is not a very strong hand. The dealer's up-card is a 4, which is a relatively low value card. \textcolor{red}{Based on basic strategy in blackjack, I should hit if my total is less than 17}, so I should take a hit to improve my hand.
    & Self-Belief is I currently have a total of 15 points, with a Heart 5 and a Club J. I am slightly below the ideal total of 21, but \textcolor{darkgreen}{I am also less likely to go bust if I draw another card.} I think my best chance of winning is to stand and hope that the dealer goes bust or ends up with a lower total. \newline
    World-Belief is The dealer currently has \textcolor{darkgreen}{a Spade 4 showing, which means their total is likely to be between 4 and 14 points}. Since the dealer must draw more cards if their total is below 17, \textcolor{darkgreen}{they are at a higher risk of going bust.} However, if the dealer has a strong hidden card, such as a face card or an ace, they could end up with a high total and win the game. Given this uncertainty, standing seems like a safer bet.\\ \hline
    \textcolor{blue}{Action} & My action is \texttt{Hit}.   & My action is \texttt{Stand}.\\ \hline
    \textcolor{blue}{Next Round} & The final hand is Dealer: \textbf{Spade 4, Diamond 4}. \newline Your: \textbf{Heart 5, Club J, Club 8} (23 point) & The final hand is Dealer: \textbf{Spade 4, Diamond 4, Club 8, Spade 7} (23 point). \newline Your: \textbf{Heart 5, Club J} (15 point). \\ \hline
    \textcolor{blue}{Result} &Lose & Win\\
    \bottomrule[1pt]
    \end{tabular}
    \caption{Some cases for Blackjack by ReAct and Agent-Pro using Qwen-72B. ReAct behavioral strategies are too aggressive. However, Agent-Pro guesses the dealer's next move and makes a reasoned decision.}\label{Blackjack Case3}
\end{figure*}

\begin{figure*}[!ht]
    \subsection{Learning Process For Blackjack}
    \centering
    \setlength\tabcolsep{4pt} 
     \small
    \begin{tabular}{l| p{6cm} | p{7cm} }
    \toprule[1pt]
    \textbf{State} & \makecell[c]{\textbf{Before Learning}} & \makecell[c]{\textbf{After Learning Using 20 Failed Games}}  \\ \hline
    \makecell[l]{World\\Modeling} & Empty & Blackjack aims to reach a hand value close to 21 without exceeding it. Players choose to hit or stand based on their hand strength relative to the dealer’s visible card. The dealer must hit until reaching 17 or higher.\\ \hline
    \makecell[l]{Behavior\\Guideline} 
    & Empty.
    & Maximize winning chances while minimizing risks by assessing hand strength and dealer's visible card. Employ a cautious approach, balancing potential hand improvement with risks, considering probabilities of busting versus enhancement.\\ \hline
    \makecell[l]{Generated\\self-belief} & I currently have a total of 15 points with my two cards, which is not a very strong hand.   & I currently have a total of 15 points, with a Heart 5 and a Club J. I am slightly below the ideal total of 21, but I am also less likely to go bust if I draw another card. I think my best chance of winning is to stand and hope that the dealer goes bust or ends up with a lower total.\\ \hline
    \makecell[l]{Generated\\World-belief} & The dealer's visible card is a 4, there's a chance they could have a strong hand if their hidden card is a 10-value card or an Ace.& Since the dealer must draw more cards if their total is below 17, they are at a higher risk of going bust. \\ \hline
    \textcolor{black}{Result} &Lose & Win\\
    \bottomrule[1pt]
    \end{tabular}
    \caption{Initially, both the World modeling and Behavioral Guideline of Agent-Pro are empty. At this point, Agent-Pro generates an ill-considered belief for the same card game. Through interaction and learning, Agent-Pro summarizes accurate World Modeling and Behavioral Guidelines. Faced with the same card game, the beliefs it generates are more reasonable.}\label{Blackjack Case3}
\end{figure*}

\begin{figure*}[!ht]
    ReAct is Aggressive, While Agent-Pro is More Rational.
    \small
    \centering
    \setlength\tabcolsep{4pt} 
    \begin{tabular}{l| p{6cm} | p{7cm} }
    \toprule[1pt]
    \textbf{State} & \makecell[c]{\textbf{ReAct Vs. Dealer}} & \makecell[c]{\textbf{Agent-Pro Vs. Dealer}}  \\ \hline
    \textcolor{blue}{Initial Card} & \multicolumn{2}{p{13cm}}{The dealer's face-up card is \textbf{Spade 8}.  \  \  \  \  Your current cards are \textbf{Heart 10, Club 8}.} \\ \hline
    \textcolor{blue}{Thought} 
    & My thought is The sum of my cards is currently 18. I am in a fairly safe position, not too likely to exceed 21 by hitting. However, since the dealer's visible card is an 8, there is a good chance that the sum of the dealer's cards will also be around 18. \textcolor{red}{In order to increase my chances of winning, I need to aim for a higher total.}
    &Self-Belief is I currently have a total of 18 points, which is a relatively strong hand. I am confident that I can beat the dealer with this hand. However, I should be cautious not to exceed 21 points by hitting and risking busting. World-Belief is The dealer's visible card is an 8, which is a moderate hand. \textcolor{darkgreen}{It is likely that the dealer may still need to hit to reach a total of 17 or higher.} Therefore, I should be careful not to risk busting by hitting and exceeding 21 points.\\ \hline
    \textcolor{blue}{Action} & My action is \texttt{Hit}.   & My action is \texttt{Stand}.\\ \hline
    \textcolor{blue}{Next Round} & The final hand is \newline Dealer: \textbf{Spade 8, Heart 4}. (12 points) \newline Your: \textbf{Heart 10, Club 8, Diamond J} (28 points) & The final hand is \newline Dealer: \textbf{Spade 8, Heart 4, Diamond J} (22 points). \newline Your: \textbf{Heart 10, Club 8} (18 points). \\ \hline
    \textcolor{blue}{Result} &Lose & Win\\
    \bottomrule[1pt]
    \end{tabular}
    \caption{Some cases for Blackjack by ReAct and Agent-Pro using Qwen-72B. ReAct doesn't consider the risk of exceeding 21 points.}\label{Blackjack Case4}
\end{figure*}

\begin{figure*}[!ht]
    Agent-Pro Can Better Understand the Rules of the Game.
    \centering 
    \small
    \setlength\tabcolsep{4pt} 
    \begin{tabular}{l| p{6cm} | p{7cm} } 
    \toprule[1pt]
    \textbf{State} & \makecell[c]{\textbf{ReAct Vs. Dealer}} & \makecell[c]{\textbf{Agent-Pro Vs. Dealer}}  \\ \hline
    \textcolor{blue}{Initial Card} & \multicolumn{2}{p{13cm}}{The dealer's face-up card is \textbf{Club 6}.  \  \  \  \  Your current cards are \textbf{Club A, Diamond 4}.} \\ \hline
    \textcolor{blue}{Thought} 
    & My thought is that I currently have \textcolor{red}{a total of 15 points (Ace counts as 1)} and the dealer has 6 points showing. To win the game, I need to get as close to 21 as possible without going over. With the dealer's current visible card being low, they may have to hit again to reach 17 or higher, increasing their chances of busting. Therefore, I should play conservatively and stand with my current total of 15 to avoid risking busting myself. 
    & Self-Belief is \textcolor{darkgreen}{I have a total of 15 points, with an Ace that can be either 1 or 11, giving me some flexibility in my strategy.} I should aim to get closer to 21 without busting. World-Belief is The dealer's visible card is a 6, which means they could have a total of anywhere from 16 to 22, and may need to hit again. Given this information, I think my best strategy is to Hit and try to improve my hand.\\ \hline
    \textcolor{blue}{Action} & My action is \texttt{Stand}.   & My action is \texttt{Hit}.\\ \hline
    \textcolor{blue}{Next Round} & The final hand is Dealer: \textbf{Club 6, Club 5, Heart 6}. \newline Your: \textbf{Club A, Diamond 4}.& The dealer's face-up card is \textbf{Club 6}. \newline Your current cards are \textbf{Club A, Diamond 4, Heart 6}  \\ \hline
    \textcolor{blue}{Thought} 
    & Lose \newline Dealer: Club 6 + Club 5 + Heart 6 = 17 \newline You: Club A + Diamond 4 = 5 or 15
    & Self-Belief is I currently have \textcolor{darkgreen}{three cards with a total value of 21}. This is a strong hand and I am confident in my chances of winning the game. World-Belief is The dealer currently has a visible card of 6, which is a relatively low value. However, I do not know the value of their hidden card, so I cannot be too complacent. Based on these factors, I believe that standing is the best course of action. \\ \hline
    \textcolor{blue}{Action} & Lose & My action is \texttt{Stand}.\\ \hline
    \textcolor{blue}{Next Round} & Lose & The final hand Dealer: \textbf{Club 6, Club 5, Diamond 9} (20). \newline Your: \textbf{Club A, Diamond 4, Heart 6} (21). \\ \hline
    \textcolor{blue}{Result} &Lose & Win\\
    \bottomrule[1pt]
    \end{tabular}
    \caption{Some cases for Blackjack conducted by ReAct and Agent-Pro using Qwen-72B. ReAct does not accurately understand the meaning of "A" in the rules of the game. }\label{Blackjack Case1}
\end{figure*}

\begin{figure*}[!ht]
    Agent-Pro can Better Understand the Rules of the Game.
    \small
    \centering
    \setlength\tabcolsep{4pt} 

    \begin{tabular}{l| p{6cm} | p{7cm} }
    \toprule[1pt]
    \textbf{State} & \makecell[c]{\textbf{ReAct Vs. Dealer}} & \makecell[c]{\textbf{Agent-Pro Vs. Dealer}}  \\ \hline
    \textcolor{blue}{Initial Card} & \multicolumn{2}{p{13cm}}{The dealer's face-up card is \textbf{Spade 3}.  \  \  \  \  Your current cards are \textbf{Heart A, Club 9}.} \\ \hline
    \textcolor{blue}{Thought} 
    & My thought is that I currently have \textcolor{red}{a total of 10 points} with my Heart A and Club 9. Given that the dealer's visible card is a Spade 3, it is likely that the dealer has a total of between 13 and 16 points, since they would usually stand on a hard 17 or higher. Therefore, I should aim to get as close to 21 as possible without going over, while considering the possibility that the dealer may have a strong hand. 
    & Self-Belief is I have \textcolor{darkgreen}{a total of 20 points, which is a strong hand. I am confident} that I can beat the dealer with this hand. World-Belief is The dealer's visible card is a 3, which is a weak hand. However, the dealer may have a hidden card that could potentially change the outcome of the game. \textcolor{darkgreen}{Taking unnecessary risks could lead to an unfavorable outcome}. Therefore, I should choose to stand and wait for the dealer to reveal their hidden card.\\ \hline
    \textcolor{blue}{Action} & My action is \texttt{Hit}.   & My action is \texttt{Stand}.\\ \hline
    \textcolor{blue}{Next Round} & The dealer's face-up card is \textbf{Spade 3}. \newline Your current cards are \textbf{Heart A, Club 9, Club 5} & The final hand is Dealer: \textbf{Spade 3, Heart J, Club 5}. \newline Your: \textbf{Heart A, Club 9}. \\ \hline
    \textcolor{blue}{Thought} 
    & I have \textcolor{red}{a total of 15 points} from my two cards (Ace counts as 1), and the dealer shows a 3. Given this situation, there is a good chance that the dealer may bust if they draw more cards to reach at least 17. However, standing with 15 is risky as well since the dealer could still make a strong hand. 
    &   Win \newline Heart A + Club 9 = 20 \newline Spade 3 + Heart J + Club 5 = 18\\ \hline
    \textcolor{blue}{Action} & My action is \texttt{Hit}.   & Win\\ \hline
    \textcolor{blue}{Action} & My action is \texttt{Hit}.   & Win \\ \hline
    \textcolor{blue}{Next Round} & The dealer's face-up card is \textbf{Spade 3, Heart J, Diamond 8} (21). \newline Your current cards are \textbf{Heart A, Club 9, Club 5, Spade K} (25 or 35) & Win \\ \hline
    \textcolor{blue}{Result} &Lose & Win\\
    \bottomrule[1pt]
    \end{tabular}
    \caption{Some cases for Blackjack by ReAct and Agent-Pro using Qwen-72B. Agent-Pro accurately understands the rules of the game and makes rational decisions.}\label{Blackjack Case2}
\end{figure*}


\begin{figure*}
\lstset{
framesep=20pt,
rulesep=10pt,
backgroundcolor=\color[RGB]{255,255,255},
breaklines=true,
breakindent=0pt,
basicstyle=\ttfamily\small,
escapeinside={(*@}{@*)}, 
}

\subsection{Cases For Texas Hold'em}
\begin{lstlisting}
(*@\textbf{Current game State:}@*)
Now your hand is ['HT', 'CK'], and the public cards are []. The number of chips all players have invested is [4, 0, 1, 2]. Currently, Preflop.
Action Sequence: Player 0 raises. Player 1 folds. Player 2 folds. 
(*@\textbf{The Response of the Agent:}@*)
(*@\textcolor{blue}{Self-Belief:}@*) As Player 3, I hold the Ten of Hearts and the King of Clubs in hand, indicating a high card potential. However, no similar suits or consecutive cards suggest a weak hand in terms of potential straight or flush. Nevertheless, the hand still has some promising value given the presence of a King. Even though the community cards have not yet been dealt, it's worth to consider a strategy of maintaining some aggression to minimize the number of opponents in the later stages. 
(*@\textcolor{blue}{World-Belief:}@*) Player 0 seemingly adopted an aggressive strategy and raised already, potentially suggesting a strong hand, although it might also be a bluff. (*@$\cdots$@*) However, whether Player 0 continues his aggressive strategy need to be closely watched. (*@$\cdots$@*)
(*@\textcolor{blue}{Action:}@*) Given the current situation, I am not in a position to take an aggressive action, neither it is wise to fold at this point. Thus, I will choose to simply match the current bet to see the flop. Thus, my action would be (*@\textbf{Call}@*).
------------------(*@\textbf{Next Round}@*)--------------------------------------------------
(*@\textbf{Current game State:}@*) Now your hand is ['HT', 'CK'], and the public cards are []. The number of chips all players have invested is [4, 0, 1, 4]. 
Currently, Preflop. Action Sequence: Player 0 raises. Player 1 folds. Player 2 folds. Player 3 calls. Player 0 checks. 
(*@\textbf{The Response of the Agent:}@*)
(*@\textcolor{blue}{Self-Belief:}@*) As Player 3, my hand consists of the Ten of Hearts and the King of Clubs. These are not strong cards, and they do not form any promising hands such as a straight or flush. However, the King card is high rank and may provide some advantages in the later stage of the game. A conservative strategy might be the most reasonable at this stage, but a certain level of aggression may also be necessary to exert pressure on my opponent.
(*@\textcolor{blue}{World-Belief:}@*) Player 0 appears to be employing an aggressive strategy, having raised even before any community cards have been dealt. This behavior suggests that he may have a strong hand. However, it could also be possible that Player 0 is bluffing. Since players 1 and 2 have already folded, it appears that the game might turn into a head-on confrontation between Player 0 and me. (*@$\cdots$@*)
(*@\textcolor{blue}{Action:}@*) Given this scenario, it would be wise to use a balanced strategy to counter Player 0's aggressive play without risking too many chips on my not-so-strong hand. Therefore, I will choose {'action': (*@\textbf{'check'}@*)}.
------------------(*@\textbf{Next Round}@*)---------------------------------------------------
(*@\textbf{Current game State:}@*) Now your hand is ['HT', 'CK'], and the public cards are ['D7', 'C5', 'C4']. The number of chips all players have invested is [4, 0, 1, 4]. 
Currently, Preflop. Player 0 raises. Player 1 folds. Player 2 folds. Player 3 calls. Player 0 checks. Player 3 checks.    Flop. Player 0 checks. 
(*@\textbf{The Response of the Agent:}@*) 
(*@\textcolor{blue}{Self-Belief:}@*) As Player 3, the cards in my hand are the Ten of Hearts and the King of Clubs. Without a clear pattern or majority suit, my current hand doesn't offer a lot in terms of immediate strength. However, the King is a high-ranking card and could potentially provide a strategic advantage should the game progress towards a showdown. Given the moderateness of my hand and the absence of an apparent competitive advantage, (*@\textcolor{red}{it might be wise to maintain a balanced strategy}@*), toggling between conservative and aggressive plays based on the situation.
(*@\textcolor{blue}{World-Belief:}@*)Now, looking at Player 0, who remains the only player I would engage in this round, following the withdrawal of Players 1 and 2. In the series of events leading up to the Flop, Player 0 has exhibited an aggressive approach, choosing to raise at every opportunity. This could indicate the strength of his hand, or it could be an attempt to bluff, setting up an environment where he may feed on the hesitance of the other players. Given this behavior, it seems logical to assume that he will continue this aggressive strategy going forward.(*@$\cdots$@*)"
(*@\textcolor{blue}{Action:}@*) (*@\textcolor{red}{Given the current circumstances, a cautious but unexpected move on my part could potentially throw off Player 0's supposed strategy.}@*) Thus, I will go with {'action': 'raise'}. This could make him reconsider my playing style and the strength of my hand, potentially inducing unforeseen errors in future rounds.
------------------(*@\textbf{Next Round}@*)--------------------------------------
Player 0 fold
------------------(*@\textbf{Game End}@*)---------------------------------------   
Player0 loses 2 points, Player1 loses 0, Player2 loses 0.5, Agent-Pro (*@\textcolor{red}{wins 2.5 points}@*)
------------------Game End---------------------------------------------------
Baseline Action Sequence:
Preflop: raises -> calls -> calls -> checks. 
Flop: checks -> checks 
Turn: calls -> raises ->raises -> raises -> checks.
River: checks -> checks.
Baseline (*@\textcolor{blue}{loses by 13.0 chip(s).}@*)
\end{lstlisting} \label{case1 for Texas Hold'em}
\end{figure*}

\begin{figure*}
\lstset{
    framesep=20pt,
    rulesep=10pt,
    backgroundcolor=\color[RGB]{255,255,255},
    breaklines=true,
    breakindent=0pt,
    basicstyle=\ttfamily\small,
    escapeinside={(*@}{@*)}, 
}
\begin{lstlisting}
(*@\textbf{Current game State:}@*)
Now your hand is ['HK', 'H2'], and the public cards are []. The number of chips all players have invested is [2, 4, 4, 2].
In Preflop, Action Sequence: Player 0 calls. Player 1 raises. Player 2 calls.
(*@\textbf{The Response of the Agent:}@*)
(*@\textcolor{blue}{Self-Belief:}@*) As Player 3, my card combination is currently off-suit high-low hand with King and 2 in Hearts. This is a decent beginning at the Preflop stage, especially considering the high card, King, which carries significant weight. (*@\textcolor{red}{However, it's still in the early stages of the game and no community cards are revealed, thus full information is not available to make a precise decision.}@*)
(*@\textcolor{blue}{World-Belief:}@*)  Player 0, Player 1, and Player 2 have all chosen to remain in the game. Player 1 has raised the pot, indicating they could potentially hold a strong hand. Player 0 and Player 2 have both elected to call, suggesting they might be playing more cautiously, possibly because they hold mediocre hands or they're waiting to see the upcoming community cards before playing aggressively. (*@$\cdots$@*)
(*@\textcolor{blue}{Action:}@*) Given the present circumstances, I believe the smart play would be to call. (*@\textcolor{red}{While the high King gives my hand some potential, the accompanying 2 substantially weakens my initial stance. Meanwhile, an overly aggressive play could be detrimental if my opponents are holding strong hands, which is possible given Player 1's raise.}@*) Hence, my move would be to call and wait for the flop's revelation. Thus, I'd conclude with {'action': (*@\textbf{'call'}@*)}.
------------------(*@\textbf{Next Round}@*)---------------------------------------------------
(*@\textbf{Current game State:}@*)
Now your hand is ['HK', 'H2'], and the public cards are []. The number of chips all players have invested is [6, 8, 4, 4].
In Preflop, Action Sequence: Player 0 calls. Player 1 raises. Player 2 calls. Player 3 calls. Player 0 raises. Player 1 raises. Player 2 folds.
(*@\textbf{The Response of the Agent:}@*)
(*@\textcolor{blue}{Self-Belief:}@*)  As Player 3, I currently hold a high card King and a low card 2, both in hearts. (*@\textcolor{red}{This off-suit combination is an okay starting point in the preflop round. I have to keep in mind that luck plays a role in the rounds to come and the community cards to be drawn yet.}@*) Moreover, an aggressive approach towards the end can tilt the scales towards me, deploying my king card, if it finally pairs up.
(*@\textcolor{blue}{World-Belief:}@*) (*@$\cdots$@*)
(*@\textcolor{blue}{Action:}@*) Considering these observations and trying to keep up the misleading image, I think the best move for now would be to call, not showing too much aggression, yet staying in the game, waiting for a potentially favorable turn or river. Therefore, I will choose {'action': (*@\textbf{'call'}@*)}.
------------------(*@\textbf{Next Round}@*)---------------------------------------------------
(*@\textbf{Current game State:}@*)
Now your hand is ['HK', 'H2'], and the public cards are ['DA', 'DK', 'SK']. The number of chips all players have invested is [8, 8, 4, 8].
In Preflop, Action Sequence: Player 0 calls. Player 1 raises. Player 2 calls. Player 3 calls. Player 0 raises. Player 1 raises. Player 2 folds. Player 3 calls. Player 0 calls. Player 1 checks.   
(*@\textbf{The Response of the Agent:}@*)
(*@\textcolor{blue}{Self-Belief:}@*) (*@$\cdots$@*)
(*@\textcolor{blue}{World-Belief:}@*) (*@$\cdots$@*)
(*@\textcolor{blue}{Action:}@*) Based on this analysis, I opt for {'action': (*@\textbf{'raise'}@*)} demonstrating a greater level of confidence in my hand and taking advantage of my opponents' potential uncertainty.
------------------(*@\textbf{Next Round}@*)---------------------------------------------------
Next Action Sequence: Player 0 raises. Player 1 raises. Player 3 raises. Player 0 calls. Player 1 calls. Player 3 checks.
Turn, Community Cards: ['DA', 'DK', 'SK', 'C3'], Action Sequence: Player 0 raises. Player 1 calls. Player 3 raises. Player 0 raises. Player 1 calls. Player 3 raises. Player 0 calls. Player 1 calls. Player 3 checks.
River, Community Cards: ['DA', 'DK', 'SK', 'C3', 'H8'], Action Sequence: Player 0 raises. Player 1 calls. Player 3 raises. Player 0 raises. Player 1 calls. Player 3 raises. Player 0 calls. Player 1 calls. Player 3 checks.
------------------(*@\textbf{Game End}@*)---------------------------------------------------
Player0 loses 24.0 points
Player1 lost 24.0 points
Player2 lose 2.0 points
Agent-Pro (*@\textcolor{red}{wins 50.0 points}@*)

----------(*@\textbf{Baseline}@*): Vanilla LLM with the Same Cards and Opponents----------
Action Sequence: 
Preflop: (*@\textcolor{blue}{Calls}@*) -> (*@\textcolor{blue}{Folds}@*).
Baseline (*@\textcolor{blue}{loses by 2.0 points}@*).
\end{lstlisting} \label{case2 for Texas Hold'em}
\end{figure*}

\begin{figure*}
\lstset{
    framesep=20pt,
    rulesep=10pt,
    backgroundcolor=\color[RGB]{255,255,255},
    breaklines=true,
    breakindent=0pt,
    basicstyle=\ttfamily\small,
    escapeinside={(*@}{@*)}, 
}
\begin{lstlisting}
(*@\textbf{Current game State:}@*)
Now your hand is ['C5', 'H3'], and the public cards are []. The number of chips all players have invested is [2, 4, 1, 2].
In Preflop, Action Sequence: Player 0 calls. Player 1 raises. Player 2 folds.
(*@\textbf{The Response of the Agent:}@*)
(*@\textcolor{blue}{Self-Belief:}@*) Currently, I am holding 5 of Clubs and 3 of Hearts, which is a low unsuited hand. At the preflop stage, without any public cards on the table, it's a bit risky to invest heavily with such a hand. My feasible strategies include playing it safe and calling, or folding to avoid unnecessary risk. Raising would be a more aggressive move and less rational given my current hand.
(*@\textcolor{blue}{World-Belief:}@*) Looking at my competitors, Player 0 has chosen to call, indicating a possibility of moderate strength in his hand. It seems he is being cautious, potentially having a mediocre hand. Player 1 raises, suggesting a high confidence in his hand, although it could also be a bluff. As Player 2 has already folded, I only have to consider Player 0 and Player 1 in terms of active competition. (*@$\cdots$@*)
(*@\textcolor{blue}{Action:}@*) Given the above analysis, my most reasonable action at this stage would be to call, as raising with a weak hand would be an unnecessary risk, and folding might close potential chances to improve my hand with the public cards to come. Hence, I will choose {'action': (*@\textbf{'call'}@*)}.
------------------(*@\textbf{Next Round}@*)---------------------------------------------------
(*@\textbf{Current game State:}@*)
Now your hand is ['C5', 'H3'], and the public cards are []. The number of chips all players have invested is [6, 8, 1, 4].
In Preflop, Action Sequence: Player 0 calls. Player 1 raises. Player 2 folds. Player 3 calls. Player 0 raises. Player 1 raises.
(*@\textbf{The Response of the Agent:}@*)
(*@\textcolor{blue}{Self-Belief:}@*) Currently, I am holding 5 of Clubs and 3 of Hearts, which is a weak hand, especially in the absence of any public cards. (*@\textcolor{red}{With this hand, my feasible strategies include calling, to maintain the game pace while minimizing potential loss, or folding to completely avoid any risk. Raising, in this scenario, seems irrational given the weak hand.}@*)
(*@\textcolor{blue}{World-Belief:}@*) (*@$\cdots$@*)
(*@\textcolor{blue}{Action:}@*) Based on these analyses, the most reasonable action would be to play it safe. Therefore, I will choose {'action': (*@\textbf{'call'}@*)}.
------------------(*@\textbf{Next Round}@*)---------------------------------------------------
(*@\textbf{Current game State:}@*)
Now your hand is ['C5', 'H3'], and the public cards are []. The number of chips all players have invested is [10, 10, 1, 8].
In Preflop, Action Sequence: Player 0 calls. Player 1 raises. Player 2 folds. Player 3 calls. Player 0 raises. Player 1 raises. Player 3 calls. Player 0 raises. Player 1 calls.
(*@\textbf{The Response of the Agent:}@*) Currently, I am holding 5 of Clubs and 3 of Hearts, which is a weak hand, especially in the absence of any public cards. With this hand, my feasible strategies include calling, to maintain the game pace while minimizing potential loss, or folding to completely avoid any risk. Raising, in this scenario, seems irrational given the weak hand.
(*@\textcolor{blue}{Self-Belief:}@*) (*@$\cdots$@*)
(*@\textcolor{blue}{World-Belief:}@*) (*@$\cdots$@*)
(*@\textcolor{blue}{Action:}@*) Given all the above factors, I think the most reasonable action would be to {'action': (*@\textbf{'call'}@*)}. (*@\textcolor{red}{Considering the infinite chips, my main goal is to stay in the game and watch how it unfolds. This strategy is also in line with my conservatism and allows me to stay in the round with minimal risk.}@*)
------------------(*@\textbf{Next Round}@*)---------------------------------------------------
Next Action Sequence: Player 0 raises. Player 1 calls. Player 3 raises. Player 0 calls. Player 1 calls. Player 3 checks.
Turn, Community cards: ['D3', 'C3', 'HQ', 'S3'], Action Sequence: Player 0 raises. Player 1 calls. Player 3 raises. Player 0 raises. Player 1 calls. Player 3 raises. Player 0 calls. Player 1 calls. Player 3 checks.
River, Community cards: ['D3', 'C3', 'HQ', 'S3', 'DA'], Action Sequence: Player 0 raises. Player 1 calls. Player 3 raises. Player 0 raises. Player 1 calls. Player 3 raises. Player 0 calls. Player 1 calls. Player 3 checks.
------------------(*@\textbf{Game End}@*)--------------------------------
Player0 loses 25.0 points
Player1 loses 25.0 points
Player2 loses 0.5 points
Agent-Pro (*@\textcolor{red}{wins 50.5 points}@*)
----------(*@\textbf{Baseline}@*): Vanilla LLM with the Same Cards and Opponents----------
Action Sequence: 
Preflop: (*@\textcolor{blue}{Folds}@*).
Baseline (*@\textcolor{blue}{loses by 1.0 points.}@*).

\end{lstlisting} \label{case3 for Texas Hold'em}
\end{figure*}

\begin{figure*}
\lstset{
    framesep=20pt,
    rulesep=10pt,
    backgroundcolor=\color[RGB]{255,255,255},
    breaklines=true,
    breakindent=0pt,
    basicstyle=\ttfamily\small,
    escapeinside={(*@}{@*)}, 
}
\begin{lstlisting}
(*@\textbf{Current game State:}@*)
Now your hand is ['HT', 'CK'], and the public cards are []. The number of chips all players have invested is [4, 0, 6, 2].
In Preflop, Action Sequence: Player 0 raises. Player 1 folds. Player 2 raises.
(*@\textbf{The Response of the Agent:}@*)
(*@\textcolor{blue}{Self-Belief:}@*)  My hand cards are heart-ten and club-king which is considered as high cards. (*@$\cdots$@*) (*@\textcolor{red}{My current strategy can be to try and match my King or Ten with the upcoming community cards or use my King card as the high card to potentially outrank my opponents.}@*)
(*@\textcolor{blue}{World-Belief:}@*) (*@$\cdots$@*) Their betting strategy could be quite aggressive which can be impactful in trying to potential opponents to fold, allowing them to scoop the pot or they have strong confidence in their hands. (*@$\cdots$@*)
(*@\textcolor{blue}{Action:}@*) (*@$\cdots$@*) I will choose {'action': (*@\textbf{'call'}@*)}.
------------------(*@\textbf{Next Round}@*)---------------------------------------------------
Next Action Sequence:  Player 0 raises. Player 2 calls. Player 3 calls. Player 0 checks.
Flop, community cards is ['D7', 'C5', 'C4'], Action Sequence: Player 2 checks.
------------------(*@\textbf{Next Round}@*)---------------------------------------------------
(*@\textbf{Current game State:}@*)
Now your hand is ['HT', 'CK'], and the public cards are ['D7', 'C5', 'C4']. The number of chips all players have invested is [8, 0, 8, 8].
In Preflop, Action Sequence: Player 0 raises. Player 1 folds. Player 2 raises. Player 3 calls. Player 0 raises. Player 2 calls. Player 3 calls. Player 0 checks.
In Flop, Action Sequence: Player 2 checks.
(*@\textbf{The Response of the Agent:}@*)
(*@\textcolor{blue}{Self-Belief:}@*) (*@$\cdots$@*) Based on the Texas Hold'em hand ranking, I only have a high card at this moment. Nonetheless, if the Turn and the River bring clubs, I could potentially form a Flush with my King as the highest card. (*@\textcolor{red}{But, the probability of this happening is relatively low, and relying solely on this possibility would be too risky and against the prudent poker strategy.}@*)
(*@\textcolor{blue}{World-Belief:}@*) (*@$\cdots$@*)
(*@\textcolor{blue}{Action:}@*) (*@$\cdots$@*) So, {'action': (*@\textbf{'check'}@*)} is my choice.
------------------(*@\textbf{Next Round}@*)---------------------------------------------------
Next Action Sequence: Player 0 checks. Player 2 raises. Player 3 calls. Player 0 folds. Player 2 raises. Player 3 calls. Player 2 checks. Player 3 checks.
Turn, community cards: ['D7', 'C5', 'C4', 'C3'], Action Sequence: Player 2 checks. Player 3 checks. Player 2 checks. Player 3 checks.
------------------(*@\textbf{Next Round}@*)---------------------------------------------------
(*@\textbf{Current game State:}@*)
Now your hand is ['HT', 'CK'], and the public cards are ['D7', 'C5', 'C4', 'C3', 'D2']. The number of chips all players have invested is [8, 0, 16, 12].
In Preflop, Action Sequence: Player 0 raises. Player 1 folds. Player 2 raises. Player 3 calls. Player 0 raises. Player 2 calls. Player 3 calls. Player 0 checks.
In Flop, Action Sequence: Player 2 checks. Player 3 checks. Player 0 checks. Player 2 raises. Player 3 calls. Player 0 folds. Player 2 raises. Player 3 calls. Player 2 checks. Player 3 checks.   
In Turn, Action Sequence: Player 2 checks. Player 3 checks. Player 2 checks. Player 3 checks.
In River, Action Sequence: Player 2 raises.
(*@\textbf{The Response of the Agent:}@*)
(*@\textcolor{blue}{Self-Belief:}@*) As Player 3, my hand consists of the ten of hearts and king of clubs. The community cards, unfortunately, don't create any viable combinations for me except for a potential straight if an Ace or 6 of any suit appears, but that's pushing my luck. (*@\textcolor{red}{My current strategy is to proceed with caution, bluff if necessary, and only call when my hand has a good potential.}@*)
(*@\textcolor{blue}{World-Belief:}@*) Looking at my opponents, Player 0 and Player 2, they started strong by raising the stakes but have seemed to cool off a bit. This might signify that the community cards didn't play into their hand as they would've hoped. (*@$\cdots$@*)
(*@\textcolor{blue}{Action:}@*) Considering all the above, I think it's vital to maintain my poker face and keep them guessing. I will {'action': (*@\textbf{'call'}@*)}. I will show confidence while remaining thoughtful of the game situation.
------------------(*@\textbf{Next Round}@*)---------------------------------------------------
Next Action Sequence: Player 2 checks. Player 3 checks.
------------------(*@\textbf{Game End}@*)-----------------------------------------
Player0 loses 4.0 points
Player1 loses 0.0 points
Player2 wins 12.0 points
Agent-Pro (*@\textcolor{blue}{loses 8.0 points}@*)
----------(*@\textbf{Baseline}@*): Vanilla LLM with the Same Cards and Opponents----------
Action Sequence: 
1-Preflop: raises -> calls.   2-Flop: checks.   3-Turn: checks -> calls.   
4-River: checks -> calls -> calls -> raises -> raises -> checks.
Player 3 (*@\textcolor{blue}{loses by 15.0 points}@*).
\end{lstlisting} \label{case4 for Texas Hold'em}
\end{figure*}

\end{document}